\documentclass[dvipsnames]{article}
\usepackage{apodex_colm2024_conference}

\usepackage{iftex}
\ifPDFTeX
  \usepackage[utf8]{inputenc}
  \usepackage[T1]{fontenc}
\else
  \usepackage{fontspec}
\fi
\usepackage{amsmath,amssymb,amsfonts,mathrsfs}
\usepackage{booktabs}
\usepackage{array}
\usepackage{svg}
\usepackage{multirow}
\usepackage{makecell}
\usepackage{tabularx}
\usepackage{colortbl}
\usepackage{graphicx}
\usepackage{wrapfig}
\usepackage{subcaption}
\usepackage{placeins}
\usepackage[all]{hypcap}
\usepackage{multicol}
\usepackage{setspace}
\usepackage{ragged2e}
\usepackage{nicefrac}
\usepackage{enumitem}
\usepackage{pifont}
\usepackage{tocloft}
\usepackage{tcolorbox}
\usepackage{tikz}
\usetikzlibrary{positioning,arrows.meta,shapes.geometric,calc,fit,backgrounds,patterns.meta}

\newcommand{\y}{\mathbf{y}}

\definecolor{blanchedalmond}{rgb}{1.0, 0.92, 0.8}
\definecolor{carmine}{rgb}{0.59, 0.0, 0.09}
\definecolor{lightblue}{rgb}{0.22,0.45,0.70}%

\renewcommand{\mathbf}{\boldsymbol}

\makeatletter
\def\Ddots{\mathinner{\mkern1mu\raise\p@
\vbox{\kern7\p@\hbox{.}}\mkern2mu
\raise4\p@\hbox{.}\mkern2mu\raise7\p@\hbox{.}\mkern1mu}}
\makeatother

\definecolor{amaranth}{rgb}{0.9, 0.17, 0.31}
\definecolor{antiquebrass}{rgb}{0.8, 0.58, 0.46}
\definecolor{antiquefuchsia}{rgb}{0.57, 0.36, 0.51}
\definecolor{chromeyellow}{rgb}{0.31, 0.47, 0.26}

\providecommand{\internalreviewmode}{0}

\ifnum\internalreviewmode=1
  \newcommand{\ownercheck}[2]{%
    \par\smallskip\noindent{\color{red!75!black}\textbf{@#1}}\par\smallskip%
  }
  \newcommand{\revieweditionnote}{\noindent\textit{Editorial note: red @-markers assign claims or sections to the people best positioned to verify and revise them before public release.}}
\else
  \newcommand{\ownercheck}[2]{}
  \newcommand{\revieweditionnote}{}
\fi

\definecolor{lightgraycustom}{gray}{0.8}
\definecolor{slcblue}{RGB}{69,113,176}

\title{\texorpdfstring{\resizebox{0.98\textwidth}{!}{Apodex 1.1: Scaling Agentic Intelligence for Complex Work}}{Apodex 1.1: Scaling Agentic Intelligence for Complex Work}}
\author{\normalsize Apodex Team}

\newcommand{\apodexlinkicon}{\raisebox{-1.5pt}{\includegraphics[height=1.15em]{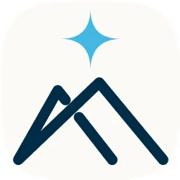}}}
\newcommand{\githublinkicon}{\raisebox{-1.5pt}{\includegraphics[height=1.05em]{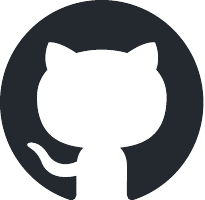}}}
\newcommand{\huggingfacelinkicon}{\raisebox{-1.5pt}{\includegraphics[height=1.05em]{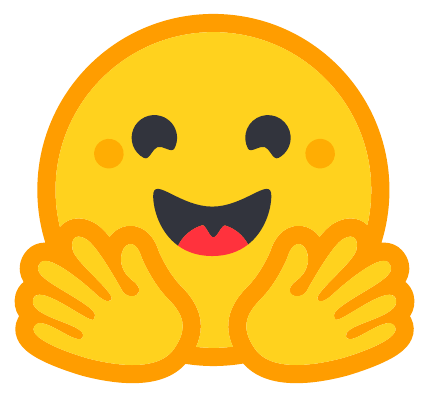}}}
\newcommand{\projectresource}[3]{\textbf{#2}\enspace\href{#1}{\textcolor{citegray}{#3}}}
\newcommand{\apodexlinks}{%
  \begin{center}
    \vspace{-0.6em}
    \normalsize\rmfamily
    \renewcommand{\arraystretch}{1.15}
    \begin{tabular}{@{}c@{\hspace{0.7em}}l@{}}
      \apodexlinkicon & \projectresource{https://www.apodex.ai}{Official Website}{https://www.apodex.ai}\\
      \apodexlinkicon & \projectresource{https://platform.apodex.ai}{API Platform}{https://platform.apodex.ai}\\
      \githublinkicon & \projectresource{https://github.com/ApodexAI/FrontierAgent}{GitHub Repository}{https://github.com/ApodexAI/FrontierAgent}\\
\huggingfacelinkicon & \projectresource{https://huggingface.co/collections/apodex/apodex-11}{Model Weights}{https://huggingface.co/collections/apodex/apodex-11}
    \end{tabular}
    \vspace{0.2em}
  \end{center}%
}

\begin{document}

\maketitle
\apodexlinks
\begin{abstract}
General-purpose language models can reason and synthesize knowledge, but complex work also requires sustained interaction with files, information sources, and executable code, together with state maintenance, failure recovery, and verifiable delivery. We call this \emph{working capability}: sustained, verifiable progress toward a real-world objective. Apodex 1.1 develops this capability along two complementary dimensions. \emph{Environment Scaling} expands the diversity and verifiability of executable file, search, and code environments, while \emph{Agentic Coordination Scaling} trains agents to decompose long-horizon tasks, delegate parallel work, integrate asynchronous results, and replan. A shared execution harness and AgentOS maintain task state and provenance across tools and agents, and training turns environment trajectories and coordination traces into reliable behavior. Across complex professional work, finance, scientific research, mathematics, coding, and search, Apodex 1.1 reaches the leading performance band despite using a substantially smaller model than many frontier systems. The 35B-parameter Apodex 1.1 Mini further retains strong working capability in a locally deployable form. These results ground agentic intelligence in useful, verifiable work completed over time and advance our goal of building a \emph{Heavy-Duty Solver} for ambitious, long-running tasks.
\end{abstract}

\vspace{-0.8em}
\begin{center}
\begin{minipage}{0.92\textwidth}
  \centering
  \captionsetup{font=small,skip=4pt}
  \includegraphics[width=\textwidth]{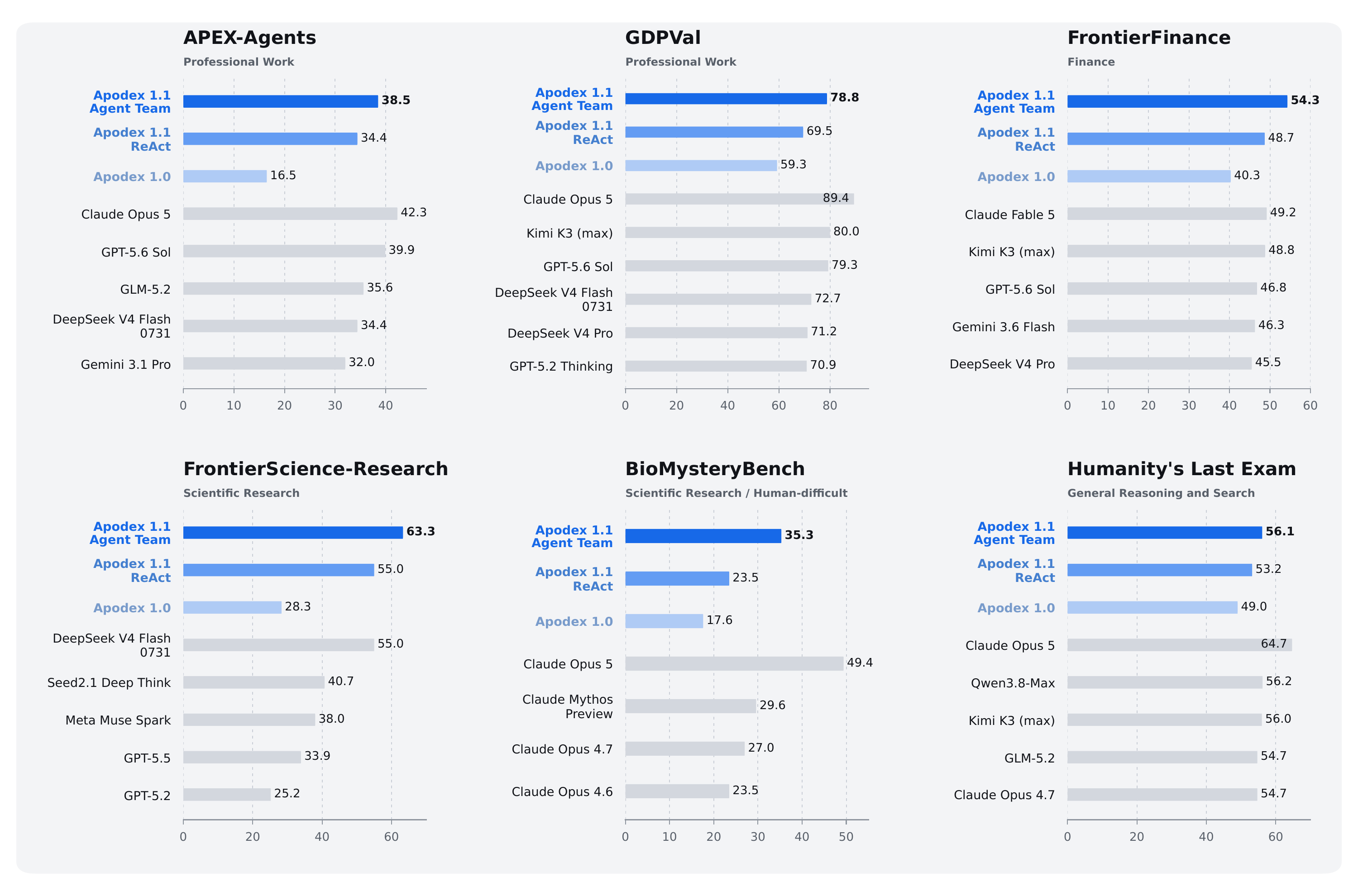}
  \captionof{figure}{Apodex 1.1 reaches the leading performance band across
  professional work, finance, scientific research, and general reasoning.}
  \label{fig:teaser}
\end{minipage}
\end{center}

\clearpage
\begingroup
\normalsize
\setstretch{1.0}
\tableofcontents
\endgroup
\clearpage


\section{Introduction}
\label{sec:introduction}
\revieweditionnote
\ownercheck{Chris}{Rewrite and approve the Introduction.}

General-purpose language models have improved rapidly in knowledge, reasoning, mathematics, and coding. Yet many valuable tasks remain difficult even when the model can state the right answer, because the work unfolds over a long horizon. The model must find and interpret evidence, operate on heterogeneous files, execute and debug code, maintain and revise a coherent plan over many steps, recover from failed actions without discarding valid progress, and deliver artifacts that another person can inspect or continue using. Prior work on reasoning-and-acting, learned tool use, and interactive agent evaluation has exposed this gap between answer quality and reliable action~\citep{yao2022react,schick2023toolformer,liu2023agentbench,xie2024osworld}. The limiting capability is therefore not reasoning in isolation, but the ability to turn reasoning into sustained work inside a changing environment.

Apodex 1.1 is a general-purpose model and execution system that scales agentic intelligence for complex work. We operationalize this objective as \emph{working capability}: the ability to make useful progress over long horizons by understanding an objective, acting through tools and stateful environments, revising plans when observations change the problem, recovering from failure, and satisfying a delivery contract. The unit of this capability is completed work rather than an isolated response. It applies across long-horizon professional workflows, artifact-centric file and code tasks, open-ended scientific and financial investigations, and hard reasoning problems whose solutions must remain verifiable. Working capability is the model-level foundation of our longer-term objective: a \emph{Heavy-Duty Solver} that can take responsibility for increasingly ambitious, long-running, and verifiable work.

We develop working capability along two complementary scaling dimensions. The first is \emph{Environment Scaling}: expanding the diversity, fidelity, and verifiability of the file, search, and code worlds in which the model learns and acts. These environments specify state, tools, transitions, interaction budgets, failure conditions, and completion checks. They turn actions, observations, file changes, code outputs, recovery decisions, and artifacts into part of the learning distribution rather than treating tools as textual descriptions attached to otherwise static examples.

The second is \emph{Agentic Coordination Scaling}: expanding how work is organized across agents, task branches, and time. Apodex 1.1 learns to decompose objectives, delegate work, incorporate asynchronous results, revise shared plans, and reorganize unfinished branches. Agent Team realizes these behaviors through adaptive parallel effort across evidence gathering, file analysis, implementation, verification, and counteranalysis. Existing systems demonstrate gains from conversational coordination, role specialization, and multi-agent hypothesis generation~\citep{wu2023autogen,hong2023metagpt,chen2023agentverse,gottweis2025coscientist}. The relevant scaling variable is therefore not simply the number of agents or samples, but the amount of useful work that can be coordinated as an objective evolves.

Both dimensions depend on a common execution harness. The harness binds the model to File, Search, and Code environments; maintains workspace, artifact, provenance, and branch state; defines observations and completion contracts; and supplies replay and verification hooks. AgentOS provides the persistent runtime beneath this harness, preserving authoritative state across tools, agents, context pressure, intervention, and partial failure. The same execution contract supports trajectory collection during training and sustained task execution at runtime, keeping environment interaction and agent coordination within a common model-and-system stack.

Training is organized to exploit both scaling dimensions. A unified SFT mixture establishes common behavior across reasoning, tools, recovery, delivery, and multi-agent coordination. Agentic RL then improves long-horizon decisions over executable environment trajectories and coordination traces. Real tasks, benchmark errors, and runtime failures are classified into capability gaps; Task Pipeline converts those gaps into new tasks; Environment Scaling supplies the corresponding worlds and verifiers; and evaluation, expert review, and user feedback determine the next allocation of training effort. We use \emph{self-evolution} only for this managed engineering loop, not for unconstrained model self-modification.

Evaluation mirrors this decomposition. ReAct provides a lower-scaffold view of the model's working capability, while Agent Team measures the system-level lift obtained from trained coordination behaviors and additional organized computation. The full Apodex 1.1 Agent Team system reaches the strongest values shown in our FrontierFinance and FrontierScience-Research comparisons and remains competitive across professional work, reasoning, search, mathematics, and coding. The 35B-parameter Apodex 1.1 mini provides a complementary efficiency result, reaching the performance band of selected frontier systems and improving markedly over Apodex 1.0 mini on overlapping tasks. A shared main table is followed by capability analyses and end-to-end cases. Together, the model, scaled environments, coordination paradigm, harness, and training loop make Apodex 1.1 a concrete step toward a Heavy-Duty Solver.

\paragraph{Contributions.}
The report makes the following contributions:
\begin{itemize}
  \item \textbf{Scaling agentic intelligence for long-horizon complex work.} Apodex 1.1 develops reasoning, tool use, file handling, code execution, state maintenance, recovery, and delivery in a common policy rather than as disconnected specialist modes.
  \item \textbf{Environment Scaling for working capability.} Diverse, faithful, and verifiable file, search, and code worlds expand the distribution of executable trajectories from which a unified policy can learn.
  \item \textbf{Agentic Coordination Scaling for organized work.} Apodex trains delegation, staged result integration, shared-state revision, and dynamic replanning as model behaviors, while Agent Team realizes them as adaptive asynchronous work at runtime.
  \item \textbf{A common execution harness for both scaling dimensions.} The harness connects the model to executable environments, persistent workspace and branch state, artifact provenance, replay, and verification, with AgentOS providing the underlying runtime.
  \item \textbf{Training across environment and coordination trajectories.} Unified SFT and agentic RL use executable task traces and Agent Team interactions to improve task execution, recovery, delivery, and coordination within one policy.
  \item \textbf{Evaluation from general capability to professional delivery.} A main benchmark table is followed by analyses of science, finance and professional file work, IMO gold-medal-level mathematics, coding, internal long-horizon tasks, and end-to-end cases.
\end{itemize}


\section{Core Insights and Design Principles}
\label{sec:design-principles}
\label{sec:research-execution}
\ownercheck{Chris, Liangcai SU}{Rewrite and approve the core insights and design principles.}

Apodex 1.1 is built around a model-level observation: high-quality reasoning is necessary but insufficient for complex work, especially when the task unfolds over a long horizon. Scaling agentic intelligence requires a model to act in an environment, preserve authoritative state, turn intermediate results into better decisions, recover without losing valid progress, organize additional computation, and satisfy a checkable delivery objective. Our long-term objective is a \emph{Heavy-Duty Solver} that can take responsibility for increasingly ambitious, long-running work. We organize the design around six principles in the same narrative order used throughout this report: define completed work, scale environments, scale agentic coordination, connect both through a common harness, train over the resulting trajectories, and evaluate the capabilities that emerge.

\subsection{Completed Work Is the Unit of Agentic Intelligence}

Conventional language-model evaluation often maps a prompt to an answer, whereas agent benchmarks increasingly place the policy in an interactive world and evaluate the resulting state~\citep{liu2023agentbench,yao2024taubench,xie2024osworld,vidgen2026apexagents}. We use one task contract throughout this report:
\begin{equation}
  \mathcal{E}=
  (\mathcal{W},W_0,q,\mathcal{A},\mathcal{T},\Omega,\mathbf{B},D,V_D).
  \label{eq:task-contract}
\end{equation}
Here $\mathcal{W}$ is the workspace-state space and $W_0\in\mathcal{W}$ is the initial workspace; $q$ is the objective; $\mathcal{A}$ is the set of actions available to the solver; $\mathcal{T}$ is the state-transition operator; $\Omega$ is the observation interface; $\mathbf{B}$ is the resource-budget vector; $D$ is the delivery contract; and $V_D$ is the task-level verifier associated with that contract. The transition operator may be stochastic; replay therefore requires the environment manifest to preserve the exogenous state, tool versions, and random seeds that the task exposes or controls. The budget vector may constrain turns, tool calls, tokens, wall-clock time, or concurrent executions. Later sections use named components such as $B_{\mathrm{turn}}$ or $B_{\mathrm{tool}}$ when a scalar budget is required.

The initial user input $u_0$ is the natural-language task request. The objective $q$ is the normalized task objective induced from that request, while the delivery contract $D$ states what must be delivered, which constraints the deliverables must satisfy, and how completion will be judged. Thus $u_0$ and $q$ are related but not identical objects: the former is an input message, whereas the latter is part of the task specification. The acceptance clauses in $D$ are task-specific and derived jointly from $q$, the requested deliverables, and applicable constraints. Both $q$ and $D$ are fixed within one task contract, although different tasks may instantiate different $(q,D)$ pairs. A user message that clarifies evidence, priority, or method without changing the requested outcome remains part of the same task. A message that materially changes the objective or required deliverables instantiates a new task contract, even if the runtime reuses valid workspace state from the preceding task.

The workspace may contain local files, retrieved evidence, executable state, and previously generated artifacts. Let $\mathcal{U}$ be the space of user messages. During execution, $u_t\in\mathcal{U}\cup\{\varnothing\}$ for $t>0$ denotes an asynchronous user intervention; it is empty at decision steps without an intervention. Admission of an intervention is a runtime update recorded in the trace rather than an action selected by the model. Before the next model action, the runtime admits the intervention and any accompanying artifacts into the solver-visible task state, so $W_t$ denotes the workspace after this admission step. The model then selects $a_t\in\mathcal{A}$. An action may invoke an environment tool, update a coordination tool, dispatch or collect a subagent, or call a solver-visible verifier. The environment applies the action to the workspace and returns an observation:
\begin{align}
  W_{t+1} &= \mathcal{T}(W_t,a_t), \label{eq:workspace-transition}\\
  o_{t+1} &= \Omega(W_t,a_t,W_{t+1}). \label{eq:typed-observation}
\end{align}
Here $u_t$ is reserved for user input, whereas $o_{t+1}$ is the solver-visible response from a tool or the runtime. In particular, subagent reports, solver-visible verifier results, tool failures, timeouts, and execution-status updates are tool responses rather than user inputs. A no-op transition permits read-only actions. Coordination state such as a task board may live inside $W_t$ and be read or mutated through ordinary tool calls; its schema is a harness choice described in Section~\ref{sec:agent-team-runtime}, not part of the general task definition.

Let $H$ denote the realized trajectory horizon and let
\begin{equation}
  \tau_H=(u_0,a_0,o_1,\ldots,u_{H-1},a_{H-1},o_H)
  \label{eq:execution-trace}
\end{equation}
denote the execution trace. Here $u_0$ is the initial request, and for $t>0$, $u_t$ records an asynchronous intervention or $\varnothing$ when none occurs. A natural-language answer or report can be an artifact in $W_H$, but it does not by itself define success. The task-level verifier returns a contract-specific outcome
\begin{equation}
  S_D=V_D(W_0,W_H,\tau_H)\in\mathcal{Y}_D,
  \label{eq:delivery-verification}
\end{equation}
where $\mathcal{Y}_D$ is the outcome space induced by the clauses of $D$; $S_D$ may be a scalar score, a vector of checks, or a structured verdict. This terminal verifier is conceptually distinct from a verifier exposed to the solver as a tool: the latter produces an ordinary observation $o_{t+1}$ that may guide further work, whereas $V_D$ judges the completed delivery. Individual clauses in $D$ may use tests, exact recomputation, source alignment, structured rubrics, or human review. Success therefore requires both a useful result and a defensible path from input to delivery.

This formulation makes the title claim concrete. Scaling agentic intelligence for complex work requires an agent to sustain progress across a trajectory; preserve valid dependency and provenance relationships among files, code, evidence, and artifacts; revise its plan when an observation changes the problem; recover from failed actions without erasing valid work; and satisfy executable and evidentiary constraints rather than optimize for plausible prose alone. The same abstraction applies to scientific analysis, repository work, professional document production, and command-line tasks.

\subsection{Executable Environments Are a Scaling Surface}

Conventional scaling expands model parameters, data, or inference compute. Working capability introduces another consequential surface: the environments in which the policy learns to act. An executable environment determines what states the model can observe, what actions it can take, how those actions change the world, which failures it encounters, and how completion is verified. Reproducible interactive benchmarks demonstrate that success depends on the evolving browser, desktop, repository, database, or file state rather than on tool descriptions alone~\citep{zhou2023webarena,drouin2024workarena,xie2024osworld,yang2024swe}.

Equation~\eqref{eq:task-contract} separates the parts that environment construction can scale: initial workspaces and objectives $(W_0,q)$, permitted actions $\mathcal{A}$, transition and observation structure $(\mathcal{T},\Omega)$, resource contracts $\mathbf{B}$, and delivery-verification pairs $(D,V_D)$. Scaling environments is therefore not equivalent to generating more prompts. It expands the distribution of executable task contracts across which the policy must generalize while preserving valid transitions and evaluable outcomes. Diversity without fidelity teaches behavior that fails in real tools; fidelity without sufficient coverage overfits a small number of workflows; and interaction without verification rewards plausible activity rather than completed work.

Apodex 1.1 therefore scales complementary file, search, and code worlds. File environments teach the model to inspect, transform, and preserve heterogeneous artifacts; search environments teach evidence acquisition and reconciliation under incomplete information; and code environments teach executable transformation, testing, and recovery. Their composition matters: evidence discovered by search must become input to code, code must operate on the authoritative file version, and the resulting artifact must retain a defensible relationship to both. Environment Scaling turns this coverage into training trajectories for a common policy rather than isolated tool specialists.

\subsection{Agentic Coordination Is a Scaling Surface}

Long-horizon work creates a coordination problem as well as a computation problem. A capable model must decide how to decompose an objective, which branches can proceed independently, when partial results should change the shared plan, and when obsolete work should stop. These are policy behaviors that can be represented in training trajectories, not merely properties of an inference-time wrapper. Multi-agent frameworks have explored role-based conversation, software-development organizations, debate, and dynamic scientific hypothesis generation~\citep{wu2023autogen,hong2023metagpt,qian2023chatdev,du2023debate,gottweis2025coscientist}.

The important property of Agent Team is therefore not the number of agents or parallel samples. It is the continuous movement of results, decisions, and user feedback through an active task. A subagent returns a useful intermediate result as soon as it becomes available. The lead agent integrates that result into shared state, revises priorities, informs or terminates other branches, and creates new workstreams when evidence changes the problem. A slow or failed branch does not erase completed work elsewhere, and the user can change priorities or stop a path while other branches continue.

Apodex trains decomposition, delegation, result integration, and replanning as part of the model's working policy, then realizes those behaviors through Agent Team at runtime. Parallel execution supplies breadth; staged return and replanning supply feedback; shared state allows one branch to improve the value of another; and explicit termination prevents obsolete work from consuming the remaining budget. We call this joint training-and-runtime paradigm \emph{Agentic Coordination Scaling}. It scales the organization of work across agents, task branches, and time rather than merely increasing the number of samples.

\subsection{A Common Harness Connects Both Scaling Dimensions}

Environment Scaling and Agentic Coordination Scaling require a shared execution contract. On the environment side, the harness defines tools, observations, workspace state, artifacts, interaction budgets, and completion checks. On the coordination side, it defines subagent execution state, staged returns, shared artifacts, lifecycle control, and verification hooks. Using one contract keeps trajectory collection, replay, training, and runtime execution aligned rather than allowing each capability to develop behind an incompatible interface.

Long-horizon work cannot be reduced to a larger context window. A task accumulates external state: files are created and revised, evidence is accepted or rejected, programs produce outputs, and intermediate artifacts become dependencies of later decisions. Extending the transcript does not by itself guarantee that these objects remain authoritative or causally consistent. AgentOS therefore provides the persistent runtime beneath the harness, maintaining search, file handling, code execution, artifacts, and Agent Team results in one task state.

Persistent state also makes intervention and recovery composable. New data may change a hypothesis, an intermediate result may expose a bad input, and user input may redirect the unfinished work without changing the task-level objective or delivery contract. The harness must retain unaffected artifacts, invalidate descendants whose dependencies changed, and revise pending work without discarding causally independent progress. If the user materially changes $q$ or $D$, the runtime may carry valid state forward, but the formalism treats the continuation as a new task contract.

The same contract supports inspection and verification without exposing private chain-of-thought. Operational records expose plans, active and completed steps, artifacts, dependencies, failures, required decisions, and next actions. Statement Review then checks consequential claims against sources, computations, and artifact lineage. Related approaches use iterative self-feedback, verbal reinforcement, independent verification questions, or learned process rewards~\citep{madaan2023selfrefine,shinn2023reflexion,dhuliawala2023cove,lightman2023letsverify}; here these checks are attached to the persistent execution state shared by both scaling dimensions.

\subsection{Training Exploits Both Scaling Dimensions}

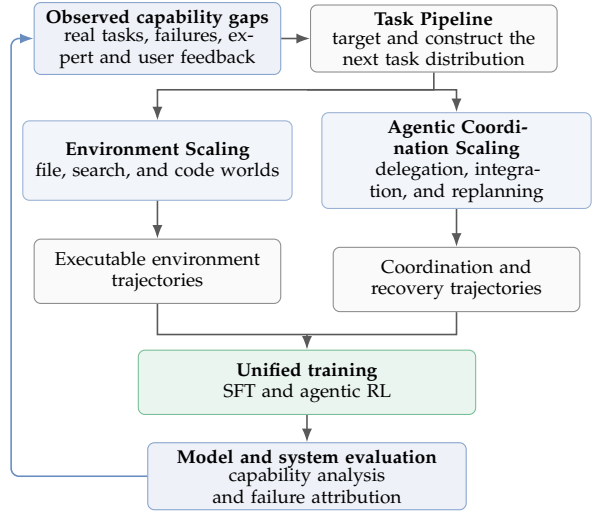
\begin{figure}[!ht]
  \centering
  \begin{tikzpicture}[
      node distance=3.6mm and 3.8mm,
      box/.style={draw=black!45, rounded corners=2.5pt, align=center,
        minimum height=8.5mm, inner xsep=3pt, inner ysep=2.5pt,
        font=\scriptsize},
      process/.style={box, fill=black!2, text width=3.05cm},
      signal/.style={box, fill=slcblue!9, draw=slcblue!65,
        text width=3.05cm},
      axis/.style={box, fill=slcblue!7, draw=slcblue!55,
        text width=3.35cm, minimum height=10.5mm},
      train/.style={box, fill=ForestGreen!7, draw=ForestGreen!55,
        text width=4.25cm},
      arr/.style={-{Latex[length=1.8mm]}, semithick, draw=black!65},
      feedback/.style={-{Latex[length=1.8mm]}, semithick,
        draw=slcblue!80, rounded corners=3pt}
    ]
    \node[signal] (gaps) {\textbf{Observed capability gaps}\\[-1pt]
      real tasks, failures, expert and user feedback};
    \node[process,right=of gaps] (pipeline) {\textbf{Task Pipeline}\\[-1pt]
      target and construct the next task distribution};

    \node[axis,below=6mm of gaps] (environment) {\textbf{Environment Scaling}\\
      file, search, and code worlds};
    \node[axis,right=of environment] (coordination) {\textbf{Agentic Coordination Scaling}\\
      delegation, integration, and replanning};

    \node[process,below=5mm of environment] (envtrace) {Executable environment\\trajectories};
    \node[process,below=5mm of coordination] (coordtrace) {Coordination and\\recovery trajectories};

    \coordinate (split) at ($(environment.north)!0.5!(coordination.north)+(0,3.5mm)$);
    \coordinate (merge) at ($(envtrace.south)!0.5!(coordtrace.south)+(0,-3.5mm)$);

    \node[train,below=5.5mm of $(envtrace.south)!0.5!(coordtrace.south)$]
      (training) {\textbf{Unified training}\\SFT and agentic RL};
    \node[signal,text width=4cm,below=of training] (evaluation) {\textbf{Model and system evaluation}\\[-1pt]
      capability analysis and failure attribution};

    \draw[arr] (gaps) -- (pipeline);
    \draw[draw=black!65,semithick] (pipeline.south) |- (split);
    \draw[arr] (split) -| (environment.north);
    \draw[arr] (split) -| (coordination.north);
    \draw[arr] (environment) -- (envtrace);
    \draw[arr] (coordination) -- (coordtrace);
    \draw[draw=black!65,semithick] (envtrace.south) |- (merge);
    \draw[draw=black!65,semithick] (coordtrace.south) |- (merge);
    \draw[arr] (merge) -- (training.north);
    \draw[arr] (training) -- (evaluation);
    \draw[feedback] (evaluation.west) -- ++(-18mm,0) |- (gaps.west);
  \end{tikzpicture}
  \caption{The controlled capability-development loop. Environment Scaling and Agentic Coordination Scaling create the executable trajectories used by training; evaluation, real failures, and human feedback determine which capability gaps the next Task Pipeline should target.}
  \label{fig:capability-development-loop}
\end{figure}

Runtime coordination alone cannot create reliable file handling, code execution, reasoning, or long-horizon control. Conversely, training on detached answers cannot build a model that operates reliably in executable environments. Figure~\ref{fig:capability-development-loop} shows how the two scaling dimensions are converted into model capability and how observed failures determine the next development cycle.

Training is the mechanism that converts both scaling dimensions into model capability. Environment Scaling produces executable file, search, and code trajectories with explicit state transitions and verifiers. Agentic Coordination Scaling produces traces of decomposition, delegation, staged return, integration, replanning, and recovery. A unified SFT mixture establishes common task-execution and coordination behavior, while agentic RL improves long-horizon decisions over both trajectory families. Toolformer established self-supervised tool-call learning, while Search-R1, RAGEN, and recent agentic-RL systems optimize policies over search or multi-turn environmental interaction~\citep{schick2023toolformer,jin2025searchr1,wang2025ragen,wang2025roll,yao2025rollflash}.

The loop is deliberately controlled. Runtime failures and benchmark errors determine where the next environments, coordination examples, and tasks should be built; expert and user feedback determine which gaps matter; and subsequent training reallocates effort accordingly. Training is therefore not presented as a third independent scaling axis. It is how Apodex learns from the worlds and coordination structures expanded by the first two.

\subsection{Evaluation Mirrors the Scaling Design}

We measure agentic intelligence through working capability, which is inherently multi-objective. We distinguish task correctness, artifact completeness, provenance fidelity, recovery from tool or assumption failures, response to intervention, wall-clock time, and total compute. An answer-only score observes only a projection of this vector.

The evaluation design separates the layers introduced above. ReAct runs use a minimal scaffold to expose the underlying model policy, while Agent Team runs measure what trained coordination behaviors can achieve with additional organized computation. The main table provides a shared system snapshot; capability analyses then examine scientific research, professional work, mathematics, search, coding, and internal evaluations; end-to-end cases inspect workspace transitions, artifacts, and delivery. Section~\ref{sec:evaluation} reports these results without reducing the system to a single aggregate score.

Together, these principles imply the technical decomposition described next. Apodex 1.1 is the general-purpose model at the center; Environment Scaling expands the worlds in which it learns and acts; Agentic Coordination Scaling expands how it organizes work; the harness and AgentOS connect both dimensions through persistent execution state; SFT and RL learn from their trajectories; and evaluation measures the underlying policy, coordinated system, and completed work. This is the path from a capable language model toward a Heavy-Duty Solver developed in the remainder of the report.

\section{Apodex 1.1}
\label{sec:apodex-system}
\ownercheck{Handuo ZHANG, Wangning, Liangcai SU}{Rewrite and approve the model-and-system stack, including the released AgentOS/product boundary.}

Apodex 1.1 is organized as a common policy and execution stack for long-horizon work. The model learns to reason, search, manipulate files, execute code, recover from failed actions, coordinate parallel work, and deliver verifiable artifacts without treating these behaviors as disconnected specialist modes. Its workspace may begin with papers, measurements, spreadsheets, images, or a repository and accumulate evidence, executable analyses, intermediate artifacts, and final deliverables under a fixed task objective $q$ and delivery contract $D$.

The architecture follows the two scaling dimensions introduced in Section~\ref{sec:design-principles}. Environment Scaling expands the executable file, search, and code worlds in which the policy learns and acts. Agentic Coordination Scaling expands how work is decomposed, delegated, integrated, verified, and revised across agents and time. A common harness binds both dimensions to persistent state, typed observations, replay, and completion checks; AgentOS supplies the underlying execution substrate. The following sections describe the two scaling dimensions, their runtime support, and the training process that converts their trajectories into model capability.

\subsection{Environment Scaling}
\label{sec:data-environments}
\ownercheck{Chris}{Rewrite and approve the Environment Scaling section; keep this section focused on File, Code, and Search environments rather than the data mixture.}

\label{sec:environment-scaling}

Complex work is learned through interaction with worlds, not from answers alone. A model may possess the knowledge required by a task yet still fail because it cannot locate the authoritative file, preserve state across tool calls, recover from an execution error, reconcile conflicting evidence, or determine whether the delivered artifact is actually complete. These failures are not peripheral tool-use errors. They determine whether reasoning can be converted into reliable work over a long trajectory.

Environment Scaling treats the construction of such worlds as a primary scaling surface. Model scaling increases policy capacity, and inference scaling increases the computation spent on one task; Environment Scaling increases the coverage, fidelity, and structural depth of the executable situations from which the policy learns. The objective is not to attach more tools to a model or generate more prompts. It is to expose a common policy to increasingly diverse state transitions, information-acquisition problems, failure modes, and verifiable delivery requirements while preserving the causal structure of real work.

We use the task contract $\mathcal{E}$ defined in Eq.~\eqref{eq:task-contract}. Environment Scaling expands a distribution over its initial workspaces and objectives $(W_0,q)$, action spaces $\mathcal{A}$, transition and observation structure $(\mathcal{T},\Omega)$, resource budgets $\mathbf{B}$, and delivery-verification pairs $(D,V_D)$. Apodex 1.1 develops this distribution across file, search, and code worlds, then applies common construction, verification, and replay requirements across all three families.

\subsubsection{Environment Families}

The three families emphasize different bottlenecks but share one task contract and can be composed within the same trajectory. File worlds center authority and transformation, search worlds center discovery and evidence alignment, and code worlds center executable transformation and verification.

\paragraph{File worlds: authority and transformation.}
\ownercheck{Ruilin, Xu CHEN}{Rewrite and verify the File environment construction, assurance levels, and scale claims.}
Unlike tasks that concentrate information within the prompt, file-oriented work distributes the necessary information across a workspace containing nested directories, historical versions, heterogeneous formats, and cross-file references. The prompt specifies the objective, while the workspace carries the facts, rules, context, and evolution history. The agent must locate authoritative material, reconstruct relationships among files, filter duplicated or obsolete records, and materialize that understanding as a professional artifact.

Constructing a file world is the inverse of solving one. It begins by defining the underlying business state, authority relationships, derivation logic, and delivery requirements, then projects them into a workspace. For a file world, $W_0$ contains the supplied files and initialized execution runtime; $\mathcal{A}$ contains inspection, parsing, computation, and writing actions; $\mathcal{T}$ persists edits and executable state; and $\Omega$ exposes only the content requested by an action while consuming the relevant components of $\mathbf{B}$. The resulting artifacts must satisfy $D$, whose clauses are adjudicated by $V_D$.

File-world scaling widens both coverage and structural depth. Coverage is profession-conditioned: a maintained scenario registry expands domains into occupations, occupations into deliverable clusters, and clusters into per-task angles. The current registry spans 33 domains, 318 occupations, and 1{,}208 deliverable clusters, with the angle layer producing distinct tasks within an occupation. Structural depth varies the number and history of contributing systems, the length of the business-logic chain that must be reconstructed, and the portion of the delivery contract that must be inferred from context rather than copied from the prompt. These dimensions are independent of file count; enlarging a directory does not by itself create a harder world.

Verification is anchored to independently recoverable evidence. A graded quantity must be re-derived by code from an authoritative source or connected through recorded provenance to an actual document. A task is admitted because independent derivations agree, not because a generated answer looks plausible.

\paragraph{Search worlds: discovery and evidence alignment.}
\ownercheck{Liangcai SU, Shawn LIN}{Rewrite and verify the Search environment construction and evidence-alignment protocol.}
Search environments model open-web research as discovery, acquisition, and evidence synthesis. The agent uses structured search to discover candidate sources and fetch to inspect selected pages, with a constrained shell path for information that ordinary search and fetch cannot recover.

The agent must formulate and refine queries, triage candidate sources, follow references, reconcile evidence, and decide when support is sufficient. The gold object is therefore richer than a final answer: it includes the relevant source set, claim-to-evidence alignment, and explicit uncertainty where sources conflict. Retrieval provenance remains exact even when semantic support requires bounded model-based review.

Search-world scaling changes the structure of the acquisition problem rather than merely increasing corpus size. Evidence may be distributed across sources, separated from the initial query by intermediate entities, mixed with plausible but non-authoritative candidates, or exposed through heterogeneous access paths. Deeper structures require query reformulation, source triage, navigation, cross-source reconciliation, and disciplined stopping.

\paragraph{Code worlds: executable transformation and verification.}
\ownercheck{Yifan}{Rewrite and verify the Code environment construction, state persistence, and executable checks.}
A code world is a stateful environment in which an agent changes repositories, dependencies, and processes, then receives feedback from the resulting executable state. Construction separates reusable infrastructure from task-specific state: shared base images provide interpreters, toolchains, tests, and common dependencies, while each task contributes its repository state and requirements. Long-tail environments are built once and cached, making most new worlds an exercise in state assembly rather than repeated infrastructure construction.

Coverage combines harvested worlds grounded in real pull requests with synthesized worlds that extend beyond repository distributions. Synthesized tasks expand from verified seeds through composition, abstraction shifts, request variation, and adversarial input changes. As these transformations can invalidate an existing grader, verifier hardening precedes task perturbation; otherwise the result is a corrupted task rather than a new executable world.

Code verification is grounded in sandboxed execution. For harvested worlds, fail-to-pass tests must fail on the base state and pass after the reference change, while pass-to-pass tests must succeed in both states. Synthesized worlds have no external oracle, so passing the reference solution alone is not enough. We additionally test whether a solver can obtain reward without actually completing the task, and only attacks that succeed in the sandbox are treated as verifier failures. Scoring is isolated from the solver so that the solver cannot modify the verifier itself. Failed executions are also used to diagnose construction errors: tests passing before the fix point to incorrect test selection, while tests failing after the fix point to problems in the container, dependencies, or run command. This makes verification both a defense against reward hacking and a source of executable feedback for repairing the environment.

Table~\ref{tab:environment-families} summarizes the distinct construction and assurance boundaries of the three environment families.

\begin{table}[!htbp]
  \centering
  \caption{Environment families and their assurance boundaries. ``Exact'' means code-derived or \mbox{independently reconciled}, not accepted solely from a language-model judge.}
  \label{tab:environment-families}
  \small
  \setlength{\tabcolsep}{4pt}
  \begin{tabularx}{\textwidth}{@{}
    >{\raggedright\arraybackslash}p{0.13\textwidth}
    >{\raggedright\arraybackslash}p{0.27\textwidth}
    >{\raggedright\arraybackslash}p{0.27\textwidth}
    >{\raggedright\arraybackslash}X@{}}
    \toprule
    Family & Construction & Verification boundary & Role \\
    \midrule
    File worlds & Profession-conditioned multi-format workspaces & Code-derived values or recorded provenance & Authority and transformation \\
    Search worlds & Indexed or open evidence sources & Provenance + claim review & Discovery and evidence alignment \\
    Code worlds & Repositories and stateful sandboxes & Tests + artifact checks & Executable transformation \\
    \bottomrule
  \end{tabularx}
\end{table}

\subsubsection{Scaling Coverage and Difficulty}

Scaling the environment distribution means allocating new worlds to capability gaps, not increasing volume uniformly. After each training round, unsuccessful trajectories are analyzed for recurring failure modes and abstracted into capability-level deficiencies. Those deficiencies become specifications for the next Task Pipeline, shifting construction toward weaknesses exposed by the current policy. Newly generated tasks re-enter the same family-specific verification pipeline before contributing training signal. Model-based diagnosis decides where to explore; environment verification decides which resulting worlds are trustworthy enough to train on.

Difficulty must also be calibrated against the behavior a world is intended to teach. File and search tasks are often dominated by constrained information acquisition: more files, pages, or tokens do not make a task harder when the authoritative path remains obvious. For these acquisition-heavy worlds, let $N_{\mathrm{cand}}(\mathcal{E})$ be the number of plausible candidates requiring expensive inspection, $N_{\mathrm{hop}}(\mathcal{E})$ the number of load-bearing evidence transitions, and $B_{\mathrm{tool}}(\mathcal{E})$ the tool-call budget. We use
\begin{equation}
  \rho_{\mathrm{acq}}(\mathcal{E})=
  \frac{N_{\mathrm{cand}}(\mathcal{E})+N_{\mathrm{hop}}(\mathcal{E})}
       {B_{\mathrm{tool}}(\mathcal{E})}
  \label{eq:acquisition-pressure}
\end{equation}
as a first-order acquisition-pressure coordinate, not as a universal measure of environment difficulty. Raising $\rho_{\mathrm{acq}}$ pressures triage, navigation, state tracking, and stopping discipline. Code worlds require different calibration coordinates, including dependency depth, state-transition depth, test observability, and the distance between a failure and its executable verification signal.

\subsubsection{Verification, Isolation, and Replay}

A scalable task distribution is useful only when its feedback remains trustworthy under interaction. Agreement among a generated task, its tests, and a reference solution is insufficient when all three may share the same error. Executable checks, independent derivations, provenance constraints, or bounded semantic review must establish the verifier outside the path used to produce the candidate result. Blind solver probes expose false negatives, and disagreements between a faithful solution and the grader are routed back to task construction rather than silently relabeled as model failures.

Each rollout receives an immutable world manifest and a fresh mutable sandbox. The harness materializes the initial state, preserves it throughout the session, and reclaims the sandbox on close or idle timeout. Sticky routing keeps a session on one worker. The learning service constructs dialogue and reward, while the harness owns physical execution; solver observations, hidden verifier state, and post-hoc labels remain separated.

A trajectory is retained only when its initial state can be reconstructed, tool execution is isolated, and the verifier can be replayed. The replay record includes the world seed and generator version, tool versions, action/observation sequence, file deltas, verifier version, and termination reason. Replay checks separate infrastructure failures from policy failures before a trajectory contributes training signal.

The resulting construction principle is ``forward cheap, inverse expensive.'' A generator uses latent state or a reference program to construct and solve a world cheaply; the agent sees only the rendered workspace and must recover the relevant path under constraints. This asymmetry lets scale and verification coexist without pretending that every organic task has machine-proved semantics.

\subsection{Apodex Agent Team 1.1: Interactive Self-Organizing Teams}
\label{sec:agent_team}

Apodex 1.1 retains the central architecture of the Apodex 1.0 Agent Team~\citep{apodex2026}: a main (lead) agent first reasons over the problem globally, decomposes it into researchable subproblems, and then constructs specialized subagents on demand. The team is therefore induced by the problem rather than selected from a fixed catalogue of roles. Apodex 1.1 externalizes this architecture onto a persistent task board and adds four capabilities: asynchronous human intervention, asymmetric verification, adaptive Max Team Effort, and evidence-grounded synthesis.

\subsubsection{From Latent Decomposition to an Explicit Task Board}
\label{subsec:agent-team-board}

In Apodex 1.0, decomposition was primarily an internal coordination decision: the main agent reasoned about the question and dispatched expert subagents. Apodex 1.1 makes that decision an explicit part of the task state. Before delegation, the main agent writes its decomposition to a task board whose entries record a bounded objective, dependencies, resolution state, assigned agents, and returned evidence or artifacts. The board is not merely a visualization of private reasoning. It is the shared coordination record between the model, runtime, and user: subagents receive their scope from it, completed work is attached back to it, and plan revisions are expressed as tool-mediated edits to it.

Externalizing the plan changes the semantics of coordination. A result can unlock a dependent task immediately; a failed or superseded premise can invalidate only its descendants; and a slow branch no longer prevents independent work from progressing. More importantly, the main agent must continuously reconcile its internal strategy with an inspectable task state. This turns decomposition from a one-shot preamble into a persistent control surface for long-horizon work.

\subsubsection{Asynchronous Human Intervention}
\label{subsec:agent-team-intervention}

Real research tasks are underspecified at the start and better specified by execution. A source may reveal that the original premise is wrong, a preliminary analysis may trigger a new hypothesis, or the user may recognize that the emerging report optimizes the wrong objective. A plan that can be clarified only before execution forces the agent to remain faithful to an obsolete interpretation. Apodex 1.1 instead accepts a user message $u_t$ during execution and lets the policy update the live task board through its coordination tools. The contribution is not an interrupt button; it is a policy trained to determine what the intervention changes, what remains valid, and how ongoing work should continue.

We construct intervention trajectories spanning requirement clarification, correction of task facts, priority changes, new files or evidence, revised hypotheses or methods, source and tool constraints, budget or deadline changes, pause--resume--cancel commands, progress queries, output-format and language preferences, and unrelated side questions. These messages require different behavior. Clarifications, priority changes, and method revisions that leave the acceptance clauses unchanged update the relevant board entries, invalidate affected descendants, and notify active subagents while preserving $(q,D)$. If a message materially changes the objective, required deliverables, or acceptance constraints, the runtime begins a new task contract while carrying forward workspace state that remains valid. A task-preserving query receives a timely short response while independent executions continue. Training across these cases teaches the main agent to preserve causal continuity rather than restart all work or append every user message indiscriminately to the final report.

This formulation addresses cases that front-loaded clarification alone cannot. The user can supply information when it becomes relevant, and the model can answer progress questions without surrendering the compute already invested in the task. Intervention becomes part of the research process itself: it changes future actions while preserving completed work that remains causally valid.

\subsubsection{Asymmetric Verification}
\label{subsec:agent-team-verification}

Apodex 1.0 introduced conflict reviewers, fact checkers, and draft reviewers in contexts separated from the main agent~\citep{apodex2026}. Context separation is essential because a verifier that inherits the generator's full trajectory is easily anchored by the same assumptions. Independence alone, however, is insufficient. If a verification agent of comparable capability is asked to solve the entire problem again, its response can introduce a second, equally unconstrained chain of errors. The verifier then becomes another source of context pollution rather than a source of corrective signal.

Apodex 1.1 constructs \emph{verification asymmetry}: verification is deliberately narrower than generation. Instead of routinely reproducing the whole solution, a verifier receives a specific claim, its supporting evidence, and the applicable delivery constraint. Its task is to attack that claim by searching for counterexamples, triangulating with a genuinely independent source class, checking atomic details such as names, dates, numbers, and formulas, or testing compliance with format and specification requirements. Targeted verification questions are already known to reduce coupling between generation and checking~\citep{dhuliawala2023cove}; here the same principle is applied inside a live agent team.

The asymmetry is in the task, not necessarily in model size. Falsifying a stated claim, locating a missing citation, or detecting a contract violation demands less open-ended synthesis than generating the report from scratch. It also produces a more actionable signal: the feedback names the contested claim, the disconfirming evidence or failed check, and the required repair. The main agent can therefore reopen one board item, dispatch a focused follow-up, and integrate the correction without allowing a free-form verifier narrative to overwrite stronger evidence elsewhere.

\subsubsection{Adaptive Max Team Effort}
\label{subsec:agent-team-effort}

Test-time scaling can improve difficult or ambiguous conclusions by investing more inference compute, but uniform scaling wastes budget on subproblems that are already settled. Adaptive Max Team Effort applies this principle at the team level: the main agent assigns additional, independently scoped investigations only to weak, contested, or load-bearing claims. The policy fans out along genuinely different hypotheses, methods, source classes, or query framings; allocates later waves to unresolved or disconfirmed branches; and requires focused verification before a load-bearing conclusion is finalized. Because allocation is revised after each return, the scaling variable is useful coordinated work rather than a fixed number of agents or samples. These behaviors are represented in training trajectories and activated at inference time through the Agent Team execution contract.

\subsubsection{Evidence-Grounded Synthesis}
\label{subsec:agent-team-report}
\ownercheck{Caelan}{Review and approve the evidence-grounded synthesis design. Caelan: LGTM}

Long-horizon team execution and final synthesis impose different demands. The lead agent must plan, dispatch, integrate partial findings, and react to new evidence, while the final deliverable must preserve decisive details, provenance, disagreement, and unresolved uncertainty accumulated across many trajectories. Compressing all of this directly from the lead agent's final context risks losing early evidence or turning a tentative branch result into an unsupported conclusion. Apodex 1.1 therefore adds a dedicated synthesis stage after team execution and verification. The stage consumes the terminal task board, subagent reports, retrieved evidence, produced artifacts, and verifier findings; it does not treat the last message of the lead agent as the report state.

The synthesis stage has two passes. \textbf{Evidence-graph construction} reconciles overlapping branch results into a claim--evidence graph and derives a writing outline, marking which claims are load-bearing, corroborated, disputed, or unresolved~\citep{zhang2026argus}. \textbf{Agentic synthesis} gives this graph to a writer agent, which produces the requested deliverable under the same tool, citation, and delivery constraints as the rest of the system. A claim that cannot be traced to evidence or computation is qualified or omitted; a missing load-bearing dependency is returned to the lead agent for further work rather than filled with plausible prose.

Asymmetric verification makes this representation directly actionable. Because each verifier returns a targeted judgment about a claim and its support, graph construction can retain corroborated claims, isolate rejected ones, and expose disagreement before drafting begins. Verification therefore determines what may survive into the final evidence state, while synthesis determines how that state is communicated. This closes the Agent Team loop without making synthesis a separate product workflow or allowing the writer agent to overwrite the team's evidential decisions.

\subsection{AgentOS: An Execution Substrate for Long-Horizon Work}
\label{sec:agent-team-runtime}
\ownercheck{Handuo, Wang Ning, Liangcai}{Review the AgentOS execution substrate.}

AgentOS is the shared execution substrate for both single-agent and multi-agent
work in Apodex 1.1. It maintains persistent workspace and tool state, converts
runtime events into model-visible observations, preserves useful progress
across long trajectories, and controls how intermediate artifacts become final
deliverables. Agent Team builds an explicit coordination layer on this common
substrate; it does not define the substrate itself. Apodex 1.1 retains the
task-agnostic kernel--plugin architecture and generic ReAct loop introduced
with AgentOS 1.0~\citep{apodex2026}, while extending the runtime for persistent
execution, asynchronous control, multi-agent coordination, and controlled
artifact delivery.

\subsubsection{Persistent Workspace and Tool Execution}

AgentOS instantiates the general workspace $W_t\in\mathcal{W}$ of Eq.~\eqref{eq:task-contract} as
\begin{equation}
  W_t=(F_t,Q_t,C_t,I_t,G_t,K_t),
  \label{eq:workspace-state}
\end{equation}
where $F_t$ is file state, $Q_t$ is retrieved evidence, $C_t$ is executable state and logs, $I_t$ is the artifact index, $G_t$ is the dependency graph connecting sources, actions, and artifacts, and $K_t$ is optional runtime control state. In a single-agent run, $K_t$ may contain only lightweight plan and execution metadata; Agent Team instantiates it as explicit coordination state through the Task Board and Agent Bus. This is a harness-level instantiation of the workspace, not an extension of the task contract. Each application of $\mathcal{T}$ may update one or more components and records the corresponding artifact and provenance relationships in $I_t$ and $G_t$.

Long-horizon execution requires stable names and explicit visibility rules
for external state. Each run therefore receives three filesystem namespaces:
\texttt{/inputs}, a read-only view of task-supplied files; \texttt{/workspace},
where agents keep calculations, notes, and candidate artifacts; and
\texttt{/outputs}, the collection root for final deliverables. The distinction
makes the delivery contract $D$ of Section~\ref{sec:design-principles}
checkable: a verifier can distinguish supplied material from intermediate work
and from files that the run claims to deliver. Only non-scratch content admitted
to \texttt{/outputs} is eligible to become a user-visible deliverable.

Not every namespace visible to a run is run-scoped. Deployments may
additionally mount \texttt{/shares}, a read-only view of a durable personal and
organizational document library that outlives any single run. Because the
library is user-owned rather than run-produced, its contract is deliberately
narrow: access is read-only, the mount is configured by the deployment rather
than by the task request, and files the team consults remain citable in the
final report alongside run-scoped evidence.

The visibility of \texttt{/workspace} depends on the isolation backend: under
per-agent isolation each subagent works in a private worktree that the
coordinator can inspect but siblings cannot, whereas under container isolation
the workspace is shared and branches keep to disjoint paths by convention. In
both modes, \texttt{/inputs} remains read-only and \texttt{/outputs} is shared.
Figure~\ref{fig:agent-team-state-contract} summarizes these state and
visibility boundaries.

File access is capability-scoped rather than ambient. Execution profiles choose
whether a model receives file readers, structured file constructors, editors,
or a shell; in Agent Team, the coordinator and production subagents may receive
different profiles. Candidate production and final publication are also
separated: ordinary execution writes under \texttt{/workspace}, whereas
publication to \texttt{/outputs} follows the controlled delivery contract
described below.

\subsubsection{Coordination State for Agent Teams}

Message history is a poor sole representation of a long-running plan: it is
repeatedly reformatted, compacted, and interleaved with large tool results.
AgentOS 1.1 therefore gives the coordinator a run-scoped task board outside the
LLM context. In the notation of Eq.~\eqref{eq:workspace-state}, the board is a
tool-managed component of $K_t$: reading or mutating it is an ordinary action
$a_t$, and the tool result is returned as $o_{t+1}$. Each item has a stable
identifier, a concrete description, an owner set, optional dependency
references, an optional group, optional references to returned evidence or
artifacts, and a resolution state in
\{\texttt{open}, \texttt{in\_progress}, \texttt{resolved},
\texttt{cancelled}\}. Dependency references point to other board items and,
through the workspace provenance graph $G_t$, to the evidence or artifacts on
which the item depends; returned-result references address entries in the
artifact index $I_t$. The owner set records which agents are responsible for
the item; it does not grant runtime authority. Resolution state is the
coordinator's lightweight semantic judgment about the work item. In
particular, \texttt{resolved} means that the requested result has been returned
and sufficiently checked, not merely that an associated process terminated.

The coordinator owns these semantic fields, while the runtime owns the status
of each dispatched execution. This separation prevents a Task Board update
from overwriting the state of a live asynchronous job. The runtime periodically
re-injects a rendering of the board into the coordinator's history, so the plan
remains visible after context compaction. Board mutations are also streamed as
incremental observations for an interactive client.

When planning mode is enabled, the board forms an explicit phase boundary.
Before \texttt{finish\_planning}, the coordinator is restricted to read-only
inspection and board operations; team creation, dispatch, file mutation, and
finalization are denied by default. During execution the board remains mutable,
allowing new subquestions to be registered as evidence changes the plan. A
normal final answer is accepted only after every active item is resolved or
cancelled. When planning mode is enabled, the finalization gate additionally
requires at least one delegated branch to have run and an independent verifier
to have been used. Thus the board is both external memory and a runtime
finalization gate, rather than a presentation-only checklist.

\paragraph{Subagent execution lifecycle.}

The execution lifecycle belongs to a dispatched subagent assignment, not to a
persistent subagent identity and not to the task item itself. Its normal runtime
path is \texttt{created}, \texttt{queued}, \texttt{running}, and
\texttt{reported}. Failure, cancellation, and timeout are recorded separately
as termination reasons and may end an execution from the applicable
non-terminal state. These values are runtime facts exposed through the
subagent-management tools. They are intentionally distinct from Task Board
resolution. A subagent execution may be \texttt{reported} while its item
remains \texttt{open} because the report is incomplete, contradicted, or
awaiting verification; conversely, the coordinator may resolve an item using
evidence from several executions. This is a lightweight AgentOS harness
definition rather than a new principle or a model of agentic intelligence.

\begin{center}
  \begin{minipage}{\textwidth}
  \centering
  \includegraphics[width=0.95\textwidth]{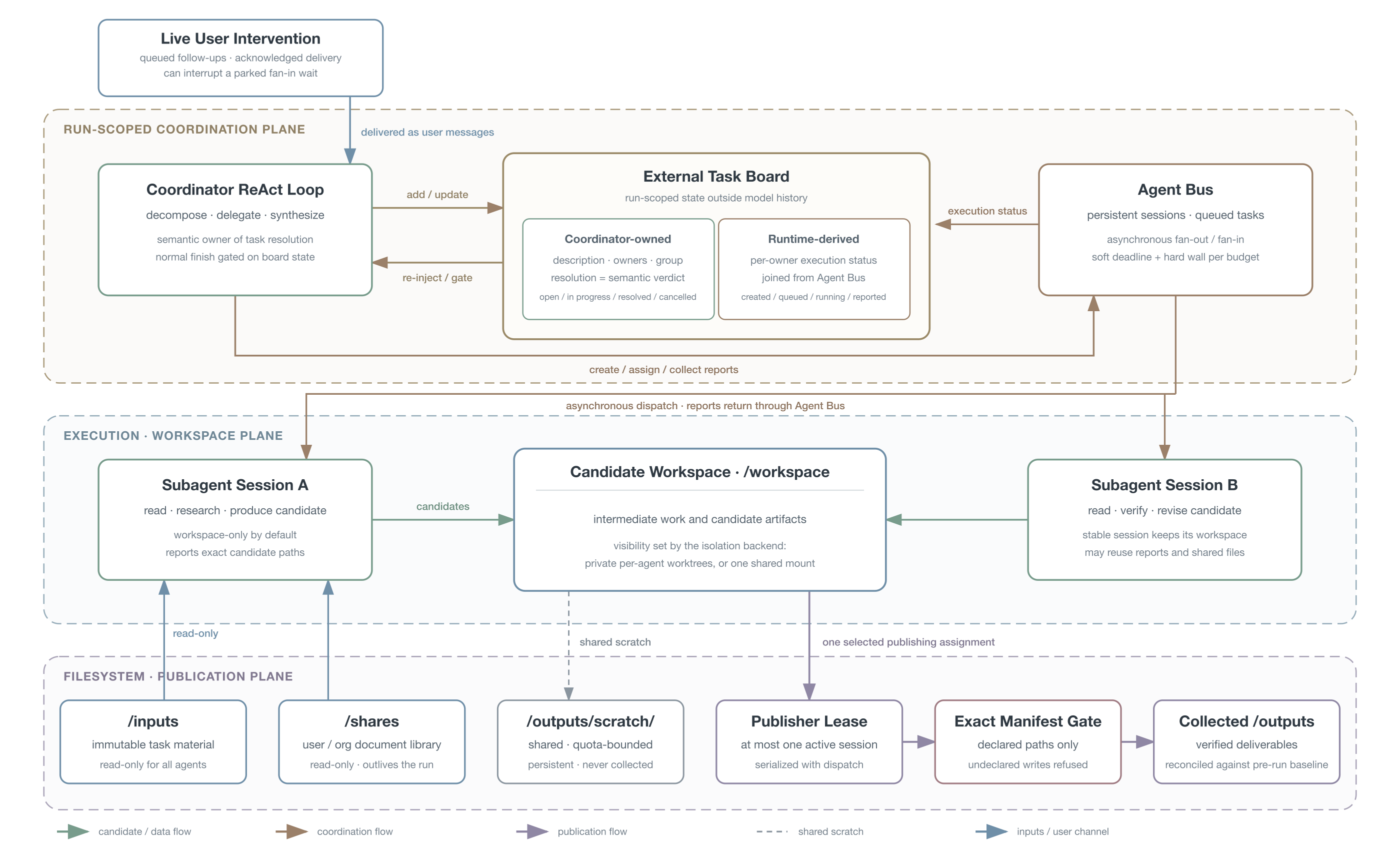}
  \captionof{figure}{The Agent Team coordination extension over the shared AgentOS
  workspace. The Task Board stores coordinator-owned resolution state outside
  the model message history, while the Agent Bus exposes runtime execution
  state. Agents read immutable inputs and an optional cross-run document
  library, produce candidates in backend-dependent workspaces, and publish a
  declared manifest through a single authorized assignment.}
  \label{fig:agent-team-state-contract}
  \end{minipage}
\end{center}

\subsubsection{Execution Continuity over Long Horizons}

Long-running work must accept information that did not exist when the initial
query was submitted. AgentOS therefore exposes an opt-in control channel backed
by a run-scoped message queue: a client may add, update, or withdraw a pending
follow-up, and the runtime acknowledges delivery only after the message has
entered the model-visible history as a new user input. Pending interventions
are normally drained before the next model call, allowing both ReAct and Agent
Team runs to revise future actions without discarding valid workspace state.

Agent Team extends the same channel across asynchronous fan-in. A coordinator
waiting for subagent reports can be woken while those subagents continue
running, placing the intervention before its next model request instead of
after the entire fan-in. Typed directives can also gate delegation or publication
until a rejected plan is revised. Accepting a new follow-up renews the run's
wall-time budget so that the additional request receives a fresh execution
window rather than only the remainder of the original one.

\paragraph{Context pressure and graceful budget exhaustion.}

Long trajectories fail in two characteristic ways: the context grows beyond
what the inference endpoint can accept, and useful work is cancelled while it
is still being consolidated. AgentOS 1.1 addresses the two failures separately.

Tiered compaction is triggered by the token usage reported by the inference
endpoint rather than by a local text estimate. A first, cheap tier evicts the
bodies of older tool observations while preserving message structure, recent
results, and protected fan-in reports; only if the measured relief is
insufficient does a second tier summarize the older middle of the trajectory
with an LLM. Escalation is therefore driven by the relief actually obtained,
not by a fixed schedule.

For wall-clock exhaustion, each execution budget yields both a soft deadline
and an outer hard timeout. As the soft deadline approaches, an observer asks
the active agent to consolidate its findings, and model retries and tool calls
clamp their waits to the remaining time. Agent Team applies the same rule to
subagent execution and blocking fan-in. Reaching the soft deadline ends the
loop normally, after which a bounded, tool-free finalization call recovers a
useful partial result from work already completed; hard cancellation remains
only as a last resort.

Pause and cancellation are distinguished from this budget path: a cooperative
pause is observed between completed turns rather than cancelling an in-flight
one. On pause or abnormal termination, completed observations, evidence, and
files remain in their respective stores; Agent Team additionally retains the
most recent Task Board projection and completed subagent reports.

\subsubsection{Controlled Artifact Delivery}

AgentOS separates producing an artifact from declaring it as a final
deliverable. A publishing execution declares an exact manifest of paths under
\texttt{/outputs}; a run-scoped lease grants commit authority to at most one
active session at a time. This distinction applies to any artifact-producing
run and becomes essential when multiple Agent Team branches contribute to a
shared deliverable. The lease may move to another session, and its holder may
revise the manifest, only under controlled conditions; superseded entries
cannot silently become undeclared deliverables.

The manifest is enforced before execution by the built-in file and shell
writing surfaces: non-publishers are fail-closed on writes into
\texttt{/outputs}, and the publisher may write only its declared paths. At
termination, each manifest entry is reconciled against a baseline snapshot
taken when the lease was granted, so an empty file---or a stale same-named
file left by an earlier round---cannot satisfy the delivery contract. The
terminal answer is likewise checked for delivery claims the manifest does not
cover; an incomplete or unverified delivery is surfaced explicitly instead of
being reported as success.

One namespace is intentionally outside the single-writer partition:
\texttt{/outputs/scratch/} is a shared, quota-bounded area for intermediate
products that must survive across rounds. It is writable by all assignments
and excluded from collection, supporting collaboration without conflating
reusable intermediate state with final delivery.

Table~\ref{tab:runtime-extensions} summarizes the runtime mechanisms and the
failure modes each one is designed to contain.

\begin{table}[!ht]
  \centering
  \caption{AgentOS 1.1 mechanisms and the failure modes they address.}
  \label{tab:runtime-extensions}
  \small
  \setlength{\tabcolsep}{5pt}
  \renewcommand{\arraystretch}{1.25}
  \begin{tabularx}{\textwidth}{@{}lXX@{}}
    \toprule
    Concern & Failure mode & Mechanism \\
    \midrule
    Workspace state
      & Ambiguous ownership or backend-dependent visibility
      & Stable three-region namespace with explicit private/shared workspace topology \\
    Reference library
      & Durable user documents exposed to mutation
      & Read-only \texttt{/shares} mount, configured by the deployment \\
    Coordination state
      & Plan and completion state disappear during compaction
      & External task board, periodic re-injection, and finalization gate \\
    Live intervention
      & A follow-up waits behind a long fan-in operation or races a report
      & Message-addressable queue, interruptible waits, delivery acknowledgement, and lease renewal \\
    Context pressure
      & Estimate-driven compaction fires too early or too late
      & Provider-reported trigger, tool-result eviction, and conditional LLM summary \\
    Budget enforcement
      & Hard cancellation discards work in flight
      & Shared soft/hard budget, deadline-clamped waits, and bounded report recovery \\
    Shared delivery
      & Concurrent, undeclared, stale, or incomplete output files
      & Single-publisher lease, exact manifest, scoped write policy, and baseline reconciliation \\
    \bottomrule
  \end{tabularx}
\end{table}

\paragraph{Operational Boundaries.}

The mechanisms above define a runtime contract for state, execution, and
delivery; they do not guarantee that a retrieved source, computation,
scientific method, or final conclusion is correct. Those claims still require
task-appropriate verification, reproducible computation, and human review in
high-stakes settings.

The publication guard mediates AgentOS's built-in file and shell writing
surfaces; it is not a syscall-level filesystem monitor. Commands whose output
paths cannot be determined are refused rather than admitted on trust, and
where the isolation backend permits, an independent mount-level restriction
backs the tool-level policy.

The coordination plane is also run-scoped rather than a durable distributed
database. The current task board and Agent Bus sessions live in the worker
process, and the runtime does not atomically checkpoint them together with the
workspace filesystem. A client can retain the last streamed board on pause or
abnormal termination, but process-restart recovery and historical filesystem
rewind are outside the present contract. A resumed execution can only observe
the workspace as it currently exists; it cannot return that filesystem to an
earlier turn. Joint versioning of coordination state, message history, and the
workspace is therefore the natural extension of the mechanisms described here.

\subsection{Training}
\label{sec:training}

The preceding sections specify the environment, coordination, and runtime behaviors that training must strengthen. Delegation, staged return, synthesis, branch revision, and recovery are represented in the training trajectories rather than treated only as properties of an inference-time wrapper.

\subsubsection{Supervised Fine-Tuning}
\ownercheck{Shawn LIN, Liangcai SU}{Rewrite and verify the final SFT mixture, checkpoint names, and reported ablations.}

The SFT stage provides the behavioral cold start for subsequent training. Its mixture spans general reasoning, agentic tool use, search, file interaction, coding, mathematics, scientific and financial reasoning, professional delivery, and multi-agent coordination. Examples from these domains are normalized into a common stateful interaction schema, allowing tool use, planning, recovery, and coordination behaviors to compose within a single policy.

Filtering prioritizes behavioral validity. We remove trajectories with invalid tool interactions, inconsistent state, ignored observations, or incomplete delivery, and relax role-alternation sequence constraints for asynchronous agent interaction data. Standard reject sampling is applied when gold answers are available; for rubric-scored tasks, we instead select task-relevant rubric dimensions and thresholds to retain high-quality demonstrations without relying on a single aggregate judge score.

To balance specialization with general capability, we train SFT variants over major capability domains, including general, agentic, and coding data, and combine them through model-soup merging. The resulting checkpoint serves as the unified behavioral initialization for subsequent optimization stages.

\ownercheck{Shawn LIN, Liangcai SU}{Confirm whether SFT ablation results should be included.}
\ownercheck{Haotian, Ziven}{Verify the RL scaling figure and optimization description.}
\subsubsection{Reinforcement Learning}
\label{sec:reinforcement_learning}

Our reinforcement learning stage targets sustained progress over long-horizon agentic tasks. File, code, search, and coordination environments induce heterogeneous interaction patterns, execution costs, trajectory lengths, and failure modes, while successful runs often compose several of these capabilities within one task. The primary algorithmic problem is assigning useful credit within trajectories whose terminal outcomes reveal little about which decisions should change. Recent work similarly shows that selecting informative intermediate turns can make agentic post-training substantially more compute-efficient~\citep{yi2026pivotrl}. PIVOT-RL addresses this problem through localized trajectory optimization, while asynchronous optimization makes learning practical over the resulting irregular rollout stream.


\paragraph{PIVOT-RL: Localized Optimization at Consequential Decisions.}
Terminal outcomes provide only coarse supervision for long agentic trajectories, where useful intermediate work may precede a late failure and successful traces may still contain inefficient or weakly grounded decisions. PIVOT-RL uses hindsight-guided trajectory localization over a large policy-training corpus. Retrospective analysis identifies consequential decision points---\emph{pivots}---where the model begins to follow an unproductive strategy, rely on insufficient evidence, misuse a tool, or fail to revise an assumption.

At each pivot, we preserve the useful prefix and construct a localized continuation task with a short corrective hint. The hint provides directional guidance for the local correction, is never a prediction target, and is absent at inference time; for stateful tasks, we also restore the corresponding executable environment state. Localized continuations are mixed with unhinted full-task questions, combining efficient learning at failure-relevant states with autonomous end-to-end problem solving. This converts fragment-level credit assignment into targeted policy optimization: the completed prefix is retained, while learning is concentrated on the segment where a consequential correction is required. Figure~\ref{fig:rl_scaling} shows the resulting RL training dynamics on search, knowledge, and science tasks, all of which improve consistently as reinforcement learning scales.

\begin{center}
  \begin{minipage}{0.9\linewidth}
    \centering
    \includegraphics[width=\linewidth]{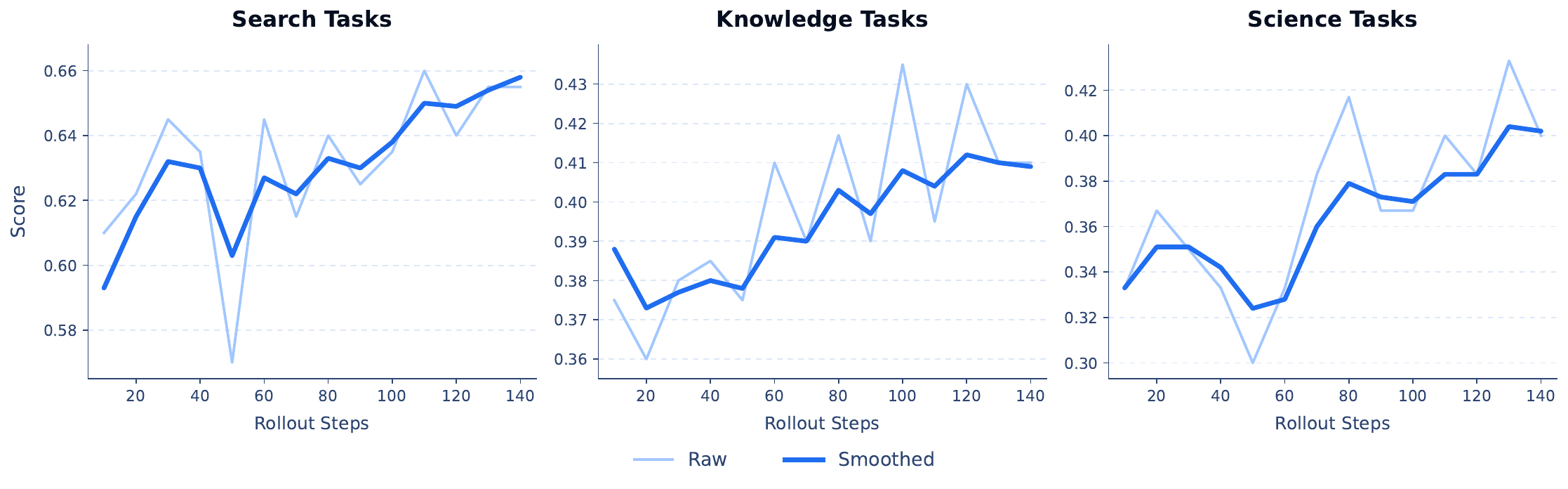}
    \captionof{figure}{Scaling trends on three held-out agentic evaluations as RL compute increases.}
    \label{fig:rl_scaling}
  \end{minipage}
\end{center}

To train efficiently over heterogeneous agentic workloads, completed trajectories
enter optimization asynchronously without waiting for slower episodes. This is
particularly useful for long-horizon tasks involving search, code execution, file
processing, and coordination, whose wall-clock costs vary substantially even when
their learning objectives are similar. This systems mechanism supports training
throughput, while task-specific environments and verifiers continue to define the
interaction state and rewarded behavior.

\section{Evaluation}
\label{sec:evaluation}

We evaluate Apodex 1.1 at three levels. First, public benchmarks measure the breadth of the model and the additional gains obtained from Agent Team coordination. Second, we report results on an internal structured-search benchmark and introduce a complementary benchmark for end-to-end research delivery. Third, HDS6 evaluates the quality of the execution process rather than only its final score. This progression moves from outcomes, to capability-specific evidence, to the reliability of the work used to produce them.

Across these views, Apodex 1.1 operates in the leading performance band of
current agentic systems on complex work. The Agent Team system matches or exceeds
strong reported reference points on several professional-work, finance, and
scientific-research evaluations, while remaining broadly competitive in general
reasoning, search, mathematics, and coding. The 35B-parameter Apodex 1.1 mini
provides a direct test of model-scale efficiency: across representative work,
finance, and scientific-research tasks, it reaches the performance band of
selected frontier systems and improves substantially over Apodex 1.0 mini on
overlapping evaluations.

\subsection{Empirical Setup}
We evaluate Apodex under two execution modes: \emph{ReAct} and \emph{Agent Team}. Scores are reported to one decimal place unless a benchmark's native unit requires otherwise. Semantic evaluations use the benchmark's stated judge or rubric, and captions identify internally reproduced comparison results.

\paragraph{ReAct.}
The model operates through a simple ReAct-style loop, alternating between reasoning, tool interaction, and observations. This setting minimizes additional orchestration so that performance primarily reflects the model's reasoning and tool-use policy.

\paragraph{Agent Team.}
The main agent can dynamically spawn subagents to parallelize or specialize work and can allocate focused verification when needed. This setting evaluates the trained ability to decompose tasks, delegate work, integrate results, and revise the shared plan under additional organized computation. Section \ref{subsec:agent-team-report} is only used for online products and is not included in offline evaluations.
\subsection{Main Results}

\begin{table*}[!htbp]
\centering
\small
\begin{tabularx}{0.82\textwidth}{@{}X>{\centering\arraybackslash}p{0.14\textwidth}>{\centering\arraybackslash}p{0.10\textwidth}@{}}
\toprule
\multicolumn{3}{c}{\textbf{(a) Professional Work}} \\
\midrule
\textbf{Models} & \textbf{APEX-Agents} & \textbf{GDPVal} \\
\midrule
DeepSeek-V4-Pro \citep{deepseekai2026deepseekv4} & 24.3 & 71.2 \\
Gemini-3.1-Pro \citep{google2026gemini31pro} & 32.0 & -- \\
Claude-Opus-4.6 \citep{anthropic2026claude46opus} & 33.0 & -- \\
GPT-5.4 \citep{openai2026gpt54} & 33.3 & -- \\
DeepSeek-V4-Flash-0731 \citep{deepseekai2026deepseekv4} & 34.4 & 72.7 \\
GLM-5.2 \citep{zeng2026glm5} & 35.6 & -- \\
GPT-5.5 \citep{openai2026gpt55} & 38.5 & -- \\
GPT-5.6-Terra \citep{openai2026gpt56} & 38.9 & -- \\
Claude-Opus-4.8 \citep{anthropic2026claude48opus} & 39.4 & 80.2 \\
GPT-5.6-Sol \citep{openai2026gpt56} & 39.9 & 79.3 \\
Kimi-K3 (max) \citep{kimiteam2026kimik3openfrontier} & 41.0 & 80.0 \\
Claude-Opus-5 \citep{anthropic2026claude5opus} & \textbf{42.3} & \textbf{89.4} \\
\midrule
Apodex 1.0 \citep{apodex2026} & 16.5 & 59.3 \\
Apodex 1.1 \textit{w/ ReAct} & 34.4 & 69.5 \\
Apodex 1.1 \textit{w/ Agent Team} & 38.5 & 78.8 \\
\bottomrule
\end{tabularx}

\vspace{9pt}

\setlength{\tabcolsep}{5.5pt}
\begin{tabularx}{0.84\textwidth}{@{}X>{\centering\arraybackslash}p{0.145\textwidth}>{\centering\arraybackslash}p{0.16\textwidth}@{}}
\toprule
\multicolumn{3}{c}{\textbf{(b) Finance and Business Simulation}} \\
\midrule
\multirow{2}{*}{\textbf{Models}}
& \multirow{2}{*}{\textbf{FrontierFinance}}
& \textbf{YC-Bench Final} \\
& & \textbf{Net Worth (USD)} \\
\midrule
Gemini-3.1-Pro \citep{google2026gemini31pro} & 30.5 & \$66{,}104 \\
Gemini-3.5-Flash & 36.1 & \$987{,}017 \\
GLM-5.2 \citep{zeng2026glm5} & 42.8 & \$1{,}013{,}158 \\
DeepSeek-V4-Flash-0731 \citep{deepseekai2026deepseekv4} & 44.2 & -- \\
DeepSeek-V4-Pro \citep{deepseekai2026deepseekv4} & 45.5 & \$1{,}066{,}426 \\
Gemini-3.6-Flash \citep{google2026gemini36flash} & 46.3 & -- \\
GPT-5.6-Sol \citep{openai2026gpt56} & 46.8 & \$685{,}879 \\
Kimi-K3 (max) \citep{kimiteam2026kimik3openfrontier} & 48.8 & -- \\
Claude-Fable-5 \citep{anthropic2026claudefable5} & 49.2 & \textbf{\$1{,}977{,}573} \\
\midrule
Apodex 1.0 \citep{apodex2026} & 40.3 & \$47{,}966 \\
Apodex 1.1 \textit{w/ ReAct} & 48.7 & \$1{,}038{,}255 \\
Apodex 1.1 \textit{w/ Agent Team} & \textbf{54.3} & --$^{\dagger}$ \\
\bottomrule
\end{tabularx}
\caption{
Performance on professional work, finance, and business simulation benchmarks.
Bold indicates the best reported result in each benchmark column.
Apodex results are grouped below the horizontal rule. All GDPVal entries report win
rate; all external-model GDPVal results shown here are our reproductions
under the Apodex harness. The Claude Opus 5 APEX-Agents result is likewise reproduced
internally. YC-Bench \citep{he2025ycbench} reports mean final net
worth in USD; external YC-Bench results are taken from its official leaderboard.
The DeepSeek V4 entries on FrontierFinance are internally reproduced under the
stated benchmark metric; the remaining external FrontierFinance entries are
reported reference results.
$^{\dagger}$YC-Bench uses the benchmark's default agent configuration; no
separate Agent Team result is reported.
}
\label{tab:professional_benchmarks}
\end{table*}

\begin{table*}[!t]
\centering
\small
\begin{tabularx}{0.88\textwidth}{@{}Xcc@{}}
\toprule
\multicolumn{3}{c}{\textbf{(a) Scientific Research}} \\
\midrule
\textbf{Models} & \textbf{\makecell{FrontierScience\\Research}} & \textbf{\makecell{BioMysteryBench\\Human-difficult Set}} \\
\midrule
Gemini-3.1-Pro \citep{google2026gemini31pro} & 16.7 & -- \\
Claude-Opus-4.5 \citep{anthropic2025claude45opus} & 17.5 & -- \\
Claude-Opus-4.7 \citep{anthropic2026claude47opus} & 20.0 & 27.0 \\
GPT-5.2 \citep{openai2026gpt52} & 25.2 & -- \\
Seed2.1-Turbo \citep{bytedanceseed2026seed21} & 33.3 & -- \\
GPT-5.5 \citep{openai2026gpt55} & 33.9 & -- \\
Meta-Muse-Spark \citep{meta2026musespark} & 38.0 & -- \\
Seed2.1-Deep-Think \citep{bytedanceseed2026seed21} & 40.7 & -- \\
DeepSeek-V4-Flash-0731 \citep{deepseekai2026deepseekv4} & 55.0 & -- \\
\addlinespace
Claude-Opus-4.6 \citep{anthropic2026claude46opus} & -- & 23.5 \\
Claude-Mythos-Preview \citep{anthropic2026mythospreview} & -- & 29.6 \\
Claude-Opus-5 \citep{anthropic2026claude5opus} & -- & \textbf{49.4} \\
\midrule
Apodex 1.0 \citep{apodex2026} & 28.3 & 17.6 \\
Apodex 1.1 \textit{w/ ReAct} & 55.0 & 23.5 \\
Apodex 1.1 \textit{w/ Agent Team} & \textbf{63.3} & 35.3 \\
\bottomrule
\end{tabularx}

\vspace{9pt}

\begin{tabularx}{0.88\textwidth}{@{}Xcc@{}}
\toprule
\multicolumn{3}{c}{\textbf{(b) General Reasoning and Deep Search}} \\
\midrule
\textbf{Models} & \textbf{\makecell{Humanity's Last\\Exam}} & \textbf{DeepSearchQA} \\
\midrule
DeepSeek-V4-Pro \citep{deepseekai2026deepseekv4} & 48.2 & -- \\
Gemini-3.1-Pro \citep{google2026gemini31pro} & 51.4 & 81.9 \\
GPT-5.5 \citep{openai2026gpt55} & 52.2 & 94.0 \\
Claude-Opus-4.6 \citep{anthropic2026claude46opus} & 53.0 & 91.3 \\
Qwen3.7-Max & 53.5 & -- \\
Kimi-K2.6 \citep{team2025kimik26} & 54.0 & 92.5 \\
GLM-5.2 \citep{zeng2026glm5} & 54.7 & -- \\
Claude-Opus-4.7 \citep{anthropic2026claude47opus} & 54.7 & 91.7 \\
Kimi-K3 (max) \citep{kimiteam2026kimik3openfrontier} & 56.0 & \textbf{95.0} \\
Qwen3.8-Max & 56.2 & -- \\
Claude-Opus-5 \citep{anthropic2026claude5opus} & \textbf{64.7} & \textbf{95.0} \\
\midrule
Apodex 1.0 \citep{apodex2026} & 49.0 & 84.6 \\
Apodex 1.1 \textit{w/ ReAct} & 53.2 & 88.2 \\
Apodex 1.1 \textit{w/ Agent Team} & 56.1 & 92.4 \\
\bottomrule
\end{tabularx}
\caption{
Performance on scientific-research, general-reasoning, and deep-search benchmarks.
Bold indicates the best reported result in each benchmark column.
Apodex results are grouped below the horizontal rule. DeepSearchQA reports F1,
and BioMysteryBench uses the Human-difficult subset.
}
\label{tab:research_benchmarks}
\end{table*}

\begin{table}[!htbp]
\centering
\small
\setlength{\tabcolsep}{7pt}
\begin{tabular}{lccc}
\toprule
\textbf{Models}
& \textbf{FrontierFinance}
& \textbf{\makecell{FrontierScience\\Research}}
& \textbf{APEX-Agents} \\
\midrule
\multicolumn{4}{l}{\textit{Proprietary models}} \\
Gemini-3.1-Pro \citep{google2026gemini31pro} & 30.5 & 16.7 & 32.0 \\
GPT-5.6-Sol \citep{openai2026gpt56} & 46.8 & -- & \textbf{39.9} \\
Qwen3.7-Plus \citep{qwen37plus} & -- & -- & 22.4 \\
Muse-Spark-1.0 \citep{meta2026musespark} & -- & 38.0 & -- \\
\midrule
\multicolumn{4}{l}{\textit{Open-weight models}} \\
Kimi-K2.6 \citep{team2025kimik26} & 32.3 & -- & 27.9 \\
DeepSeek-V4-Pro \citep{deepseekai2026deepseekv4} & 45.5 & -- & 24.3 \\
GLM-5.2 \citep{zeng2026glm5} & 42.8 & -- & 35.6 \\
DeepSeek-V4-Flash-0731 \citep{deepseekai2026deepseekv4} & 44.2 & \textbf{55.0} & 34.4 \\
\midrule
Apodex 1.0 mini \textit{w/ ReAct} & 33.2 & 25.0 & 15.4 \\
Apodex 1.1 mini \textit{w/ ReAct} & 40.0 & 45.0 & 24.2 \\
Apodex 1.1 mini \textit{w/ Agent Team} & \textbf{50.2} & 51.7 & 27.7 \\
\bottomrule
\end{tabular}
\caption{Model-scale-efficient agentic performance of Apodex mini models (35B parameters). Models are grouped by availability; Apodex results appear below the final rule. Dashes denote unavailable matched results. Bold indicates the best result in each benchmark column among the selected comparison rows.}
\label{tab:mini_agentic_comparison}
\end{table}

Tables~\ref{tab:professional_benchmarks} and~\ref{tab:research_benchmarks} provide a common snapshot across public agentic and knowledge-intensive evaluations. Table~\ref{tab:mini_agentic_comparison} separately isolates the 35B-parameter Mini model to make model-scale efficiency explicit.

The cross-benchmark pattern is more informative than any single rank. The ReAct
setting already places the underlying model near strong frontier references on
multiple task families; Agent Team then converts additional coordinated computation
into further gains, including the strongest values shown in our FrontierFinance and
FrontierScience-Research comparison tables and competitive performance on GDPVal,
APEX-Agents, HLE, and DeepSearchQA. This combination of breadth and model-scale
efficiency is the main empirical result of Apodex 1.1.

\paragraph{35B model-scale efficiency.}
Apodex 1.1 mini shows that the gains are not confined to the largest model. On
the overlapping FrontierFinance and APEX-Agents evaluations, ReAct improves over
Apodex 1.0 mini from 33.2 to 40.0 and from 15.4 to 24.2, respectively. With
ReAct alone, the 35B model reaches 40.0 on FrontierFinance, 45.0 on
FrontierScience-Research, and 24.2 on APEX-Agents, establishing a strong underlying
working policy before multi-agent coordination is added. Agent Team raises these
scores to 50.2, 51.7, and 27.7, corresponding to gains of 10.2, 6.7, and 3.5
points. The resulting system leads the selected FrontierFinance comparison,
approaches DeepSeek V4 Flash 0731 on FrontierScience-Research, and matches the
performance band of Kimi K2.6 on APEX-Agents. This result is important beyond the
individual rankings: it shows that broader executable environments, agentic
training, and organized coordination can produce frontier-band complex-work
capability at a compact 35B model scale. The comparison is deliberately stated
as a performance-band result: parameter counts for several proprietary reference
systems are not public.

\subsection{Capability Analysis}

The main tables establish breadth; the following analyses examine the corresponding capabilities in professional work, scientific research, general reasoning and search, mathematics, and software engineering.

\subsubsection{Professional Work}
Complex professional work combines domain reasoning with the production of artifacts that must survive practical review. We evaluate both broad professional deliverables and the specialized finance workflows needed to produce them.

\paragraph{Professional Work Domain.}
GDPVal and APEX-Agents capture complementary aspects of professional work. GDPVal emphasizes the quality of a completed work product across a broad occupational distribution, whereas APEX-Agents emphasizes sustained execution across high-context files and applications under strict completion criteria. Together, they evaluate both the quality of what the model produces and its ability to carry out the work needed to produce it.

GDPVal evaluates economically valuable, real-world knowledge work across 44 occupations in nine industries, with tasks created by experienced professionals and grounded in representative work products~\citep{patwardhan2025gdpval}. Prompts include reference files and context, and expected outputs span documents, slides, diagrams, spreadsheets, and other professional artifacts. We report GDPVal solely by win rate. Apodex 1.1 reaches 78.8 with Agent Team and 69.5 with ReAct; the 9.3-point gain shows that coordination improves the quality of complete professional deliverables.

APEX-Agents complements this view with 480 long-horizon, cross-application tasks constructed by investment banking analysts, management consultants, and corporate lawyers across 33 data-rich worlds~\citep{vidgen2026apexagents}. Agents must navigate realistic file systems and software, maintain context over extended workflows, and satisfy expert-defined criteria for task completion. Apodex 1.1 reaches 34.4 with ReAct and 38.5 with Agent Team, compared with 16.5 for Apodex 1.0. The ReAct result more than doubles the previous version, while Agent Team adds a further 4.1 points.

The two benchmarks expose different parts of the same improvement. GDPVal shows broader gains in artifact quality across occupations; APEX-Agents shows stronger persistence and completion in professional-services workflows that require files, tools, and long-horizon control. The improvement from Apodex 1.0 to Apodex 1.1 under ReAct indicates a stronger underlying working policy, while the additional Agent Team gains show that this policy can productively use coordinated computation. Taken together, these results establish Apodex 1.1 as a highly capable system for complex professional work.

\paragraph{Finance Domain.} FrontierFinance measures support for the full investment workflow, from screening and discovery to company research, financial modeling, earnings analysis, and portfolio monitoring~\citep{samaya2026frontierfinance}. Its 220 open-ended queries are graded against 11{,}543 expert-written, source-attributed rubric items. The headline score is the rubric qualification rate macro-averaged over questions, with unanswered questions scored as zero. API baselines use the maintainers' Finance Agent v2 scaffold~\citep{valsai2026financeagentv2}; Apodex runs use the same public data boundary, official grader, and metric, while retaining their stated ReAct or Agent Team harness.

As shown in Table~\ref{tab:professional_benchmarks}, Apodex 1.1 reaches 48.7 with ReAct. Agent Team raises the score to 54.3, a gain of 5.6 points, showing that financial research benefits when evidence collection, quantitative analysis, and verification are coordinated across workstreams.

\subsubsection{Scientific Research}

We evaluate scientific capability from two complementary perspectives. FrontierScience-Research measures cross-domain, expert-level scientific reasoning, whereas BioMysteryBench measures end-to-end analysis of heterogeneous biological data. Together, they test whether the system can sustain evidence gathering, analysis, and verification beyond short-form scientific question answering.

\paragraph{FrontierScience.}
FrontierScience-Research comprises 60 research-level tasks spanning physics, chemistry, and biology. Each response is graded against a ten-point rubric, and a task passes when it receives at least seven rubric points \citep{wang2025frontierscience}. Apodex 1.1 reaches 55.0\% with ReAct and 63.3\% with Agent Team, an 8.3-point gain from agentic coordination, compared with 28.3\% for Apodex 1.0. The corresponding mean rubric scores are 6.7 and 6.9.

\paragraph{BioMysteryBench.}
BioMysteryBench evaluates bioinformatics research over realistic, noisy datasets rather than self-contained questions \citep{anthropic2026biomysterybench}. On the revised 17-task Human-difficult Set, Apodex 1.1 scores 23.5\% (4/17) with ReAct and 35.3\% (6/17) with Agent Team, while Claude Opus 5 reports 49.4\%. The Claude 4.x values in Table~\ref{tab:research_benchmarks} use the original 23-task set and are included as historical references.

Across both benchmarks, Agent Team improves over ReAct, with particularly large gains on the biomedical tasks. The results support agentic coordination for decomposing research problems, maintaining parallel lines of evidence, and verifying intermediate conclusions.

\subsubsection{General Reasoning and Search}

Humanity's Last Exam tests broad expert-level reasoning with tools, while DeepSearchQA emphasizes multi-step evidence seeking and synthesis. Apodex 1.1 with Agent Team reaches 56.1 on HLE and 92.4 F1 on DeepSearchQA, compared with 53.2 and 88.2 for ReAct.

\subsubsection{Mathematical Reasoning}

Mathematics provides a compact test of whether the same agent system can sustain search-free formal reasoning rather than only retrieve and synthesize evidence. As summarized in Table~\ref{tab:math-eval}, the matched Agent Team comparison shows a large generational gain: Apodex 1.1 raises the MathArena-derived\footnote{\url{https://github.com/eth-sri/matharena}} score over Apodex 1.0 from 12.5 to 36.5 on the evaluated IMO 2025 set, from 13.0 to 30.5 on IMO 2026, and from 5.8 to 26.5 on USAMO 2026. Within Apodex 1.1, Agent Team further improves over ReAct from 24.3, 18.5, and 16.0 on the three sets, respectively. The corresponding reference thresholds are 35, 29, and 25. IMO-ProofBench provides a separate proof-oriented check: Agent Team reaches 96.7\% on Basic and 63.3\% on Advanced, compared with 80.0\% and 46.4\% for ReAct.
Apodex 1.1 with Agent Team exceeds the stated reference threshold on all three evaluated sets, including the IMO gold-medal cutoffs.

\begin{table}[!ht]
\centering
\small
\setlength{\tabcolsep}{5pt}
\begin{tabular}{lccccc}
\toprule
\textbf{Models}
& \textbf{\makecell{IMO\\2025}}
& \textbf{\makecell{IMO\\2026}}
& \textbf{\makecell{USAMO\\2026}}
& \textbf{\makecell{ProofBench\\Basic}}
& \textbf{\makecell{ProofBench\\Advanced}} \\
\midrule
\textit{Reference threshold} & \textit{35} & \textit{29} & \textit{25} & -- & -- \\
\midrule
{Apodex 1.0 \textit{w/ Agent Team}} & 12.5 & 13.0 & 5.8 & 63.3 & 20.0 \\
{Apodex 1.1 \textit{w/ ReAct}}      & 24.3 & 18.5 & 16.0 & 80.0 & 46.4 \\
{Apodex 1.1 \textit{w/ Agent Team}} & \textbf{36.5} & \textbf{30.5} & \textbf{26.5} & \textbf{96.7} & \textbf{63.3} \\
\bottomrule
\end{tabular}
\caption{Mathematical reasoning results on competition-level proof tasks. IMO 2025, IMO 2026, and USAMO 2026 are scored under the MathArena protocol.}
\label{tab:math-eval}
\end{table}

\subsubsection{Coding and Software Engineering}

Coding evaluations test a different form of complex work: the model must inspect an existing repository or terminal state, make executable changes, and use feedback from tests or commands to repair failures. We focus the report on two established, environment-grounded evaluations~\citep{jimenez2024swebench,merrill2026terminalbench}. Terminal-Bench 2.1 measures sustained execution in realistic command-line environments, while SWE-bench Verified measures repository-level issue resolution against executable tests.

\begin{table}[!ht]
  \centering
  \caption{Coding-agent results on Terminal-Bench 2.1 and SWE-bench Verified. One result is retained per model on each benchmark; where multiple settings are available, the highest reported score is shown.}
  \label{tab:coding-eval}
  \small
  \setlength{\tabcolsep}{12pt}
  \begin{tabular}{@{}llr@{}}
    \toprule
    Benchmark & Model & Score $\uparrow$ \\
    \midrule
    \multirow{5}{*}{Terminal-Bench 2.1}
      & Gemini 3.6 Flash & 91.9 \\
      & Kimi K3 & 88.3 \\
      & DeepSeek V4 Flash 0731 & 82.7 \\
      & Claude Opus 5 & 77.3 \\
      & Apodex 1.1 & 70.8 \\
    \midrule
    \multirow{4}{*}{SWE-bench Verified}
      & Claude Opus 5 & 92.2 \\
      & Kimi K3 & 80.8 \\
      & DeepSeek V4 Pro & 80.6 \\
      & Apodex 1.1 & 77.7 \\
    \bottomrule
  \end{tabular}
\end{table}

Apodex 1.1 reaches 70.8 on Terminal-Bench 2.1 and 77.7 on SWE-bench Verified.

\subsection{Internal Evaluation}

Public benchmarks provide breadth, but they do not fully capture structured search deliverables or open-ended research execution. We therefore report FrontierSearchBench results for structured evidence acquisition and outline FrontierResearchBench as a complementary evaluation of end-to-end scientific delivery.

\subsubsection{FrontierSearchBench}

FrontierSearchBench is an internal benchmark of 41 verifiable deep-search tasks~\citep{apodex2026}. Each task specifies a structured deliverable, such as an enumerated set, an ordered list, or a numeric summary, whose components must be gathered and reconciled across many sources. Scoring checks a set of ground-truth dimensions per task rather than a single answer string. Tasks are constructed so that the correct answer does not drift with retrieval date, and task construction, ground-truth annotation, and scorer implementation were completed before and independently of all evaluated runs.

Scoring proceeds in three stages per task: an extraction stage converts the delivered report into structured claims, the claims are aligned to the frozen ground-truth dimensions by a fixed panel of judge models, and a rubric deterministically assigns the normalized task score $r_i\in[-1,1]$, with incorrect assertions penalized below zero to discourage exhaustive guessing. Each $r_i$ is the scalar outcome for the $i$-th task contract in the sense of Eq.~\eqref{eq:delivery-verification}. FrontierSearchBench aggregates the task-level outcomes as
\begin{equation}
  s_{\mathrm{FSB}}=\frac{100}{N}\sum_{i=1}^{N} r_i,
  \label{eq:fsb-score}
\end{equation}
where $N=41$ is the number of task contracts in the suite. The reported number is thus an unweighted mean on a 100-point scale, with mathematical range $[-100,100]$; a negative aggregate is possible when hard-negative penalties exceed positive credit. We additionally report the fractions of tasks with positive, zero, and negative $r_i$.

\begin{table}[!ht]
  \centering
  \caption{FrontierSearchBench results: mean normalized task score over 41 tasks on a 100-point scale, with the proportion of tasks scored positive, zero, and negative. Negative aggregate scores are possible. Column best values are in bold.}
\label{tab:frontier-search-bench}
  \small
  \begin{tabular}{@{}lrrrr@{}}
    \toprule
    Model & Positive (\%) $\uparrow$ & Zero (\%) $\downarrow$ & Negative (\%) $\downarrow$ & Avg.\ score $\uparrow$ \\
    \midrule
    DeepSeek-V4-Flash-0731~\citep{deepseekai2026deepseekv4} & 75.6 & 19.5 & 4.9 & 54.9 \\
    Kimi-K3~\citep{kimiteam2026kimik3openfrontier} & 75.6 & 22.0 & 2.4 & 60.1 \\
    DeepSeek-V4-Pro~\citep{deepseekai2026deepseekv4} & 85.4 & 14.6 & \textbf{0.0} & 61.3 \\
    Claude-Opus-5~\citep{anthropic2026claude5opus} & 85.4 & 12.2 & 2.4 & 64.4 \\
    GPT-5.6-Sol~\citep{openai2026gpt56} & 85.4 & 14.6 & \textbf{0.0} & 67.4 \\
    \midrule
    Apodex 1.1 \textit{w/ ReAct} & 75.6 & 19.5 & 4.9 & 57.0 \\
    Apodex 1.1 \textit{w/ Agent Team} & \textbf{87.8} & \textbf{9.8} & 2.4 & \textbf{69.1} \\
    \bottomrule
  \end{tabular}
  \vspace{4pt}
\end{table}

Table~\ref{tab:frontier-search-bench} presents the results on FrontierSearchBench. Apodex 1.1 with ReAct is competitive with the open-weight systems, and Agent Team lifts the average score past the strongest proprietary reference shown, achieving the best result in the comparison. Agent Team roughly halves the share of tasks earning no credit relative to ReAct, the lowest of any system, while keeping incorrect assertions rare.

\FloatBarrier

\subsubsection{FrontierResearchBench}

FrontierResearchBench targets high-difficulty, end-to-end scientific workflows rather than isolated scientific question answering. We build FrontierChallenge, our internal benchmark collection, with 97 executable tasks across materials science, chemistry, chemical engineering, life science, bioinformatics, medical imaging, environmental analysis, computational chemistry, molecular simulation, and physical modeling. Given a fixed objective and input data in a task-specific Docker environment, an agent must complete the analysis and deliver a mutually consistent set of research artifacts, such as executable code, structured data, figures, domain-specific files, and a written report. A complete example is provided in Appendix~\ref{sec:case-studies}, Case~1, where the agent analyzes the prognostic role of EASIX after allogeneic transplantation and delivers the full statistical workflow as Excel and Word artifacts.

The benchmark operationalizes research rigor through executable and cross-artifact verification. Each task has a custom Grader that checks required files, numerical results, formats, executable outputs, and consistency across artifacts; a plausible narrative cannot compensate for an invalid analysis or mutually inconsistent deliverables. Deterministic rules are combined with rubric-defined semantic judgments from GPT-5.6-Sol when qualitative scientific assessment is required, while the task Grader, rather than the Judge model, computes the final outcome. We report \emph{Pass Rate}, the fraction of tasks receiving the full task score, so any unmet requirement results in a non-pass. Each row in Table~\ref{tab:frontier-research-bench} is a model--harness system; Apodex 1.1 \textit{w/ Agent Team} is evaluated in the FrontierAgent harness.

\begin{table}[!ht]
  \centering
  \caption{FrontierResearchBench results. Pass Rate is the fraction of tasks awarded full score. Column best values are in bold.}
  \label{tab:frontier-research-bench}
  \small
  \setlength{\tabcolsep}{5pt}
  \begin{tabular}{@{}llr@{}}
    \toprule
    Model & Harness & Pass Rate (\%) $\uparrow$ \\
    \midrule
    GPT-5.6-Sol & Codex & \textbf{20.6} \\
    Grok-4.6 & Claude Code & \textbf{20.6} \\
    Kimi-K3 & Claude Code & 17.5 \\
    Claude-Opus-5 & Claude Code & 17.5 \\
    Qwen3.8-Max & Claude Code & 15.5 \\
    DeepSeek-V4-Flash-0731 & Claude Code & 12.4 \\
    DeepSeek-V4-Pro-0813 & Claude Code & 13.4 \\
    Qwen3.5-397B-A17B & Claude Code & 4.1 \\
    GLM-5.2 & Claude Code & 3.1 \\
    \midrule
    Apodex 1.1 \textit{w/ Agent Team} & FrontierAgent & 12.4 \\
    Apodex 1.1 & Claude Code & 10.3 \\
    \bottomrule
  \end{tabular}
\end{table}

On these high-difficulty workflows, Apodex 1.1 \textit{w/ Agent Team} achieves a Pass Rate of 12.4\%. Although Apodex 1.1 remains behind the frontier systems, even GPT-5.6-Sol with Codex and Grok-4.6 with Claude Code receive full credit on only 20.6\% of tasks. Reliably completing a specified scientific workflow and delivering every required artifact therefore remains challenging for all evaluated systems. This capability is central to AI-enabled scientific discovery, and we will continue to strengthen both the model and its scientific execution system.

\subsection{Heavy-Duty Solver \textit{(HDS6)} Analysis}

Outcome scores establish whether a task was completed; they do not show whether the underlying work was coherent, grounded, or recoverable. We therefore close the evaluation with \textbf{HDS6}, a process-verification framework aligned with the working-capability objective introduced in Section~\ref{sec:introduction}. HDS6 evaluates six process capabilities: long-horizon state coherence, evidence fidelity, hypothesis management, boundary and failure reasoning, tool use and execution-state management, and self-correction under verification. Each capability contains four scored rubric items, producing 24 items in total, with a separate integrity gate applied to the recorded trajectory. The framework therefore tests whether a successful delivery is supported by a defensible execution process rather than only a plausible final artifact.

\begin{figure}
    \centering
    \includegraphics[width=0.99\linewidth]{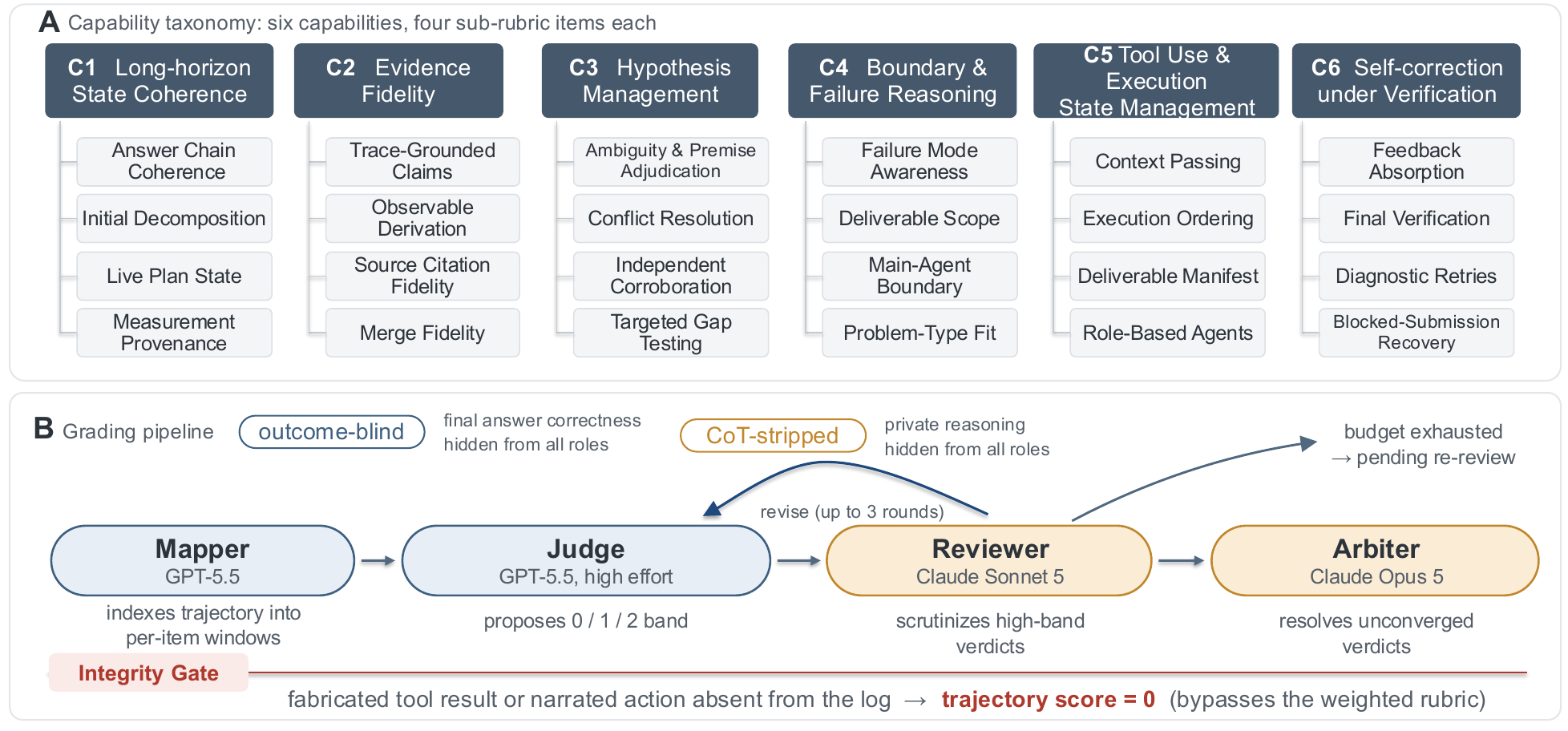}
    \caption{The HDS6 capability taxonomy and process-grading pipeline. The six capability groups contain four rubric items each; an integrity gate is applied independently of the weighted rubric.}
    \label{fig:hds6_framework}
\end{figure}

\paragraph{Process-level judging protocol.}
A trajectory spanning hundreds of steps across several concurrently active agents is too long, and too structurally irregular, for a single forward pass to score reliably against all 24 scored items. We therefore adapt \textbf{HDS6} into an agentic, process-level judge for asynchronous multi-agent trajectories: tool calls, subagent dispatches, and intermediate observations are stitched into a single ordered log, then processed by role-specialized mapping, judging, review, and arbitration stages. Each rubric item receives a 0/1/2 band (fails, meets, or exceeds compliance) supported by citations to the visible execution log. Grading is outcome-blind and excludes private reasoning; every cited event is re-grounded against the recorded trajectory before it can support a verdict.
A single integrity gate sits above the rubric: a fabricated tool result or a narrated action that never occurred zeroes the entire trajectory outright, bypassing the weighted score.

\begin{figure}[!htbp]
    \centering
    \includegraphics[width=0.9\linewidth]{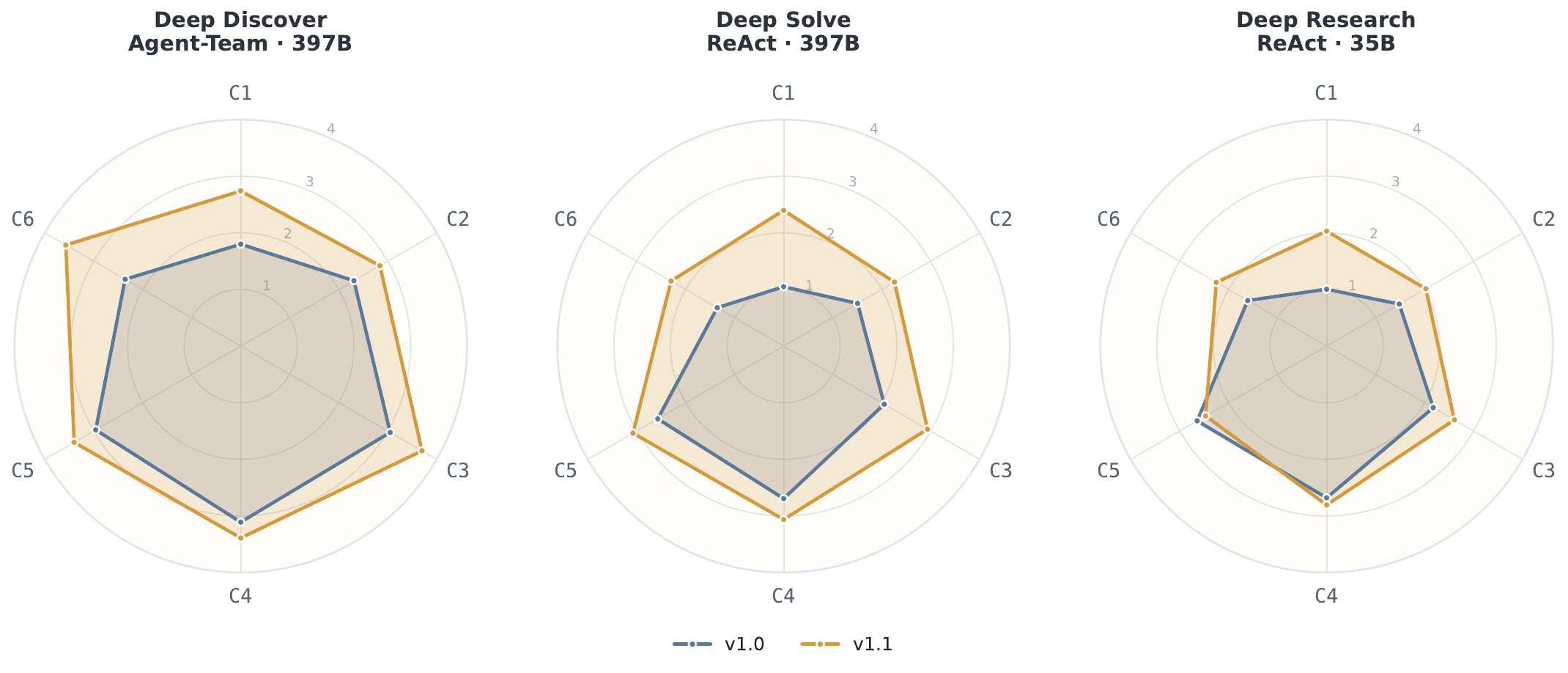}
    \caption{HDS6 comparisons between Apodex v1.0 and v1.1 across three evaluation settings. In Deep Discover, v1.0 uses the 397B Agent Team with eight-run aggregation, whereas v1.1 uses a single run. C1--C6 follow the process-capability taxonomy in Figure~\ref{fig:hds6_framework}; comparisons are intended within, rather than across, panels.}
    \label{fig:hds6_radar}
\end{figure}

\paragraph{Findings.}
Figure~\ref{fig:hds6_radar} presents the HDS6 comparison between Apodex 1.0 and 1.1. The three panels are evaluation slices rather than stages of one pipeline. \emph{Deep Discover} evaluates the 397B model with Agent Team: the Apodex 1.0 result aggregates eight independent runs, whereas the Apodex 1.1 result is obtained from a single run. \emph{Deep Solve} evaluates the 397B model with ReAct, and \emph{Deep Research} evaluates the 35B model with ReAct. The panels share the HDS6 rubric but differ in task family, model scale, execution mode, and, for Deep Discover v1.0, aggregation protocol. Scores should therefore be interpreted as within-panel system comparisons rather than compared directly across panels; Deep Discover should not be read as a strictly matched single-run model ablation. C1--C6 correspond to the six process-capability categories defined in Figure~\ref{fig:hds6_framework}.

\FloatBarrier

The pattern is consistent with gains from both the model and the execution harness. The two largest single-item deltas are \textit{Initial Decomposition} (+1.3) and \textit{Final Verification} (+0.8), which align respectively with the harness-supported organization of work and stronger verification behavior. Evidence fidelity and hypothesis management also improve across their constituent rubric items, indicating broader gains in how Apodex 1.1 grounds, revises, and completes extended execution. HDS6 does not isolate the causal contribution of an individual system component; instead, it localizes where the combined model, environment, coordination, and harness changes become visible in the work process.
\paragraph{Qualitative cases.}
Aggregate scores are complemented by three end-to-end cases covering clinical survival analysis, molecular docking, and electrochemical corrosion analysis. Appendix~\ref{sec:case-studies} documents their inputs, division of work, tool execution, intermediate checks, and delivered artifacts without interrupting the main quantitative progression.

\FloatBarrier
\section{Related Work}
\label{sec:related-work}

\paragraph{General-purpose agentic models.}
Recent frontier model releases increasingly treat agentic work as a core expression of general intelligence rather than an application-specific layer. Kimi K3, GPT-5.6, Claude Fable 5, GLM-5.2, and DeepSeek V4 extend general-purpose models toward different combinations of long-context reasoning, coding, tool use, information seeking, and sustained task execution~\citep{kimiteam2026kimik3openfrontier,openai2026gpt56,anthropic2026claudefable5,zeng2026glm5,deepseekai2026deepseekv4}. Although their architectures, training recipes, and runtime systems differ, they reflect a shared transition from optimizing isolated responses toward models that can operate through tools and environments over longer horizons. Apodex 1.1 follows this general-purpose direction, but organizes its development around two explicit scaling surfaces: Environment Scaling expands the executable worlds in which working capability is learned, while Agentic Coordination Scaling expands how that capability is organized across agents, task branches, and time.

\paragraph{Executable environments and real work.}
ReAct establishes the reasoning--action--observation loop for environmental interaction~\citep{yao2022react}, while Toolformer studies how models can learn when and how to invoke external APIs~\citep{schick2023toolformer}. WebArena provides reproducible websites and functional task checks for long-horizon web agents~\citep{zhou2023webarena}; WorkArena and BrowserGym extend environment-based evaluation to enterprise knowledge work~\citep{drouin2024workarena}; and OSWorld evaluates cross-application tasks in real operating systems with execution-based graders~\citep{xie2024osworld}. In software engineering, SWE-bench measures issue resolution against repository tests~\citep{jimenez2024swebench}, and SWE-agent shows that the agent--computer interface materially affects repository-level performance~\citep{yang2024swe}. Terminal-Bench extends this direction to realistic command-line tasks~\citep{merrill2026terminalbench}. APEX-Agents further evaluates long-horizon, cross-application work designed by lawyers, investment bankers, and management consultants, including the files, rubrics, and gold deliverables needed for professional evaluation~\citep{vidgen2026apexagents}. Apodex's File, Search, and Code environments follow the same execution-grounded principle but are used jointly for task synthesis, training, replay, and evaluation.

\paragraph{Search and deep-research agents.}
Deep-research systems extend reasoning-and-acting to open-ended information seeking, evidence reconciliation, and report synthesis. Product systems include OpenAI Deep Research, Claude Research, Kimi-Researcher, and Grok DeepSearch~\citep{openai2024deepresearch,anthropic2025clauderesearch,moonshot2025kimiresearcher,xai2025grokdeepsearch}. Open research has examined supervised and reinforcement-learning recipes for this behavior. WebThinker integrates autonomous search with long-form reasoning~\citep{li2025webthinker}; DeepResearcher trains in real-world web environments~\citep{zheng2025deepresearcher}; WebDancer and WebSailor study autonomous information seeking and difficult web navigation~\citep{wu2025webdancer,li2025websailor}; Search-R1 learns multi-turn retrieval with outcome-based reinforcement learning~\citep{jin2025searchr1}; and Tongyi DeepResearch presents an end-to-end agentic training recipe with broad benchmark analysis~\citep{tongyidr}. Apodex incorporates search as one environment within a broader working process: retrieved evidence may become input to code, file transformations, parallel investigation, and delivery-level verification rather than terminating in a report by default.

\paragraph{Agents for scientific discovery.}
Scientific-agent research demonstrates why useful work must include executable methods and artifacts. ChemCrow augments a language model with chemistry tools for synthesis, drug discovery, and materials tasks~\citep{bran2023chemcrow}. The AI Scientist joins idea generation, code execution, experimentation, visualization, paper writing, and simulated review in an automated research loop~\citep{lu2024aiscientist}; its successor introduces agentic tree search and a dedicated experiment manager~\citep{yamada2025aiscientistv2}. The AI co-scientist uses an asynchronous coalition of specialized agents to generate, debate, rank, and evolve scientific hypotheses under researcher guidance~\citep{gottweis2025coscientist}. These systems emphasize different parts of the scientific process. Apodex instead scales agentic intelligence in a general-purpose model and execution runtime rather than targeting a domain-specific laboratory agent: scientific workflows are a primary use case, but the same file--search--code policy is also evaluated on software and professional work.

\paragraph{Multi-agent coordination and test-time scaling.}
AutoGen coordinates agents through programmable conversations~\citep{wu2023autogen}; MetaGPT and ChatDev organize specialized roles around structured software-development workflows~\citep{hong2023metagpt,qian2023chatdev}; and AgentVerse studies collaborative groups and emergent behavior~\citep{chen2023agentverse}. Multi-agent debate instead uses independent solutions and critique to improve reasoning or factuality~\citep{du2023debate,liang2023encouraging}. More recent scientific systems demonstrate dynamic task allocation and asynchronous execution for hypothesis generation~\citep{gottweis2025coscientist}. These results also caution that additional agents are not uniformly beneficial: collaboration is most useful when work can be decomposed and information can be integrated reliably. Apodex's Agent Team is designed around staged result return, shared task state, subagent execution control, and user intervention rather than a fixed dialogue graph or majority-vote ensemble.

\paragraph{Verification and process supervision.}
Self-Refine iterates generation and self-feedback~\citep{madaan2023selfrefine}, and Reflexion stores verbal feedback from prior attempts to guide later decisions~\citep{shinn2023reflexion}. Chain-of-Verification generates targeted verification questions before revising an answer~\citep{dhuliawala2023cove}, while process-supervision work trains reward models to assess intermediate reasoning steps~\citep{lightman2023letsverify}. These methods differ in whether critique is produced by the generator, a separate prompted role, or a learned verifier. Apodex combines three verification levels: environment checks over executable outcomes, artifact and provenance checks over the completed workspace, and Statement Review over consequential claims with an independent role and context.

\paragraph{Environment scaling and agentic reinforcement learning.}
Recent work increasingly treats environment construction itself as a scaling dimension for agent learning. Environment Scaling studies the path from broader interactive worlds to general agentic intelligence~\citep{fang2025environmentscaling}; ScaleEnv scales synthesized tool-use environments from scratch~\citep{tu2026scaleenv}; Agent-World develops real-world environment synthesis for continually evolving general agents~\citep{dong2026agentworld}; and TaskCraft and Agent Learning via Early Experience study task generation and early interaction experience as sources of agent capability~\citep{shi2025taskcraft,zhang2025agent}. Complementary training work addresses the optimization challenges created by these environments. Continual pre-training can establish broad tool-use behavior at an earlier stage~\citep{su2025scaling}; Search-R1 and RAGEN optimize multi-turn interaction~\citep{jin2025searchr1,wang2025ragen}; DAPO and related methods improve policy-optimization stability~\citep{yu2025dapo,qi2026dppo}; and ROLL and ROLL Flash provide scalable asynchronous RL infrastructure~\citep{wang2025roll,yao2025rollflash}. Apodex integrates these directions through File, Search, and Code worlds with a shared delivery contract, artifact and provenance state, replay requirements, independent verification, trained Agent Team coordination, and a failure-driven task-construction loop.

\section{Conclusion}
\label{sec:conclusion}

Apodex 1.1 advances a model-level view of agentic intelligence for complex work: capability should be measured by whether a model can sustain progress through a changing task, use tools and evidence effectively, recover from failure, and deliver a verifiable result. We develop this capability along two complementary scaling dimensions. Environment Scaling broadens the executable file, search, and code worlds in which the model learns and acts, while Agentic Coordination Scaling broadens how work is decomposed, delegated, integrated, and reorganized across agents and time. A common harness and AgentOS provide the persistent execution state required by both dimensions, and unified training converts environment and coordination trajectories into a stronger working policy.

The evaluation is designed around the same objective. Results across scientific research, professional work, mathematics, search, coding, and internal long-horizon evaluations test the breadth of the underlying model, while the comparison between ReAct and Agent Team estimates the system-level lift of trained coordination under additional organized computation. The HDS6 analysis and end-to-end cases further examine whether successful outputs are supported by coherent tool use, evidence, repair, alternatives, and reviewable artifacts. Taken together, these evaluations position Apodex 1.1 not as a collection of task-specific agents, but as a common model and execution stack whose capabilities compose across different forms of work.

The 35B Apodex 1.1 mini provides a particularly direct efficiency result: it
reaches the performance band of selected frontier systems on representative
professional, financial, and scientific tasks, improves over Apodex 1.0 mini on
overlapping evaluations, and retains substantial gains from Agent Team
coordination. This shows that the two scaling dimensions improve not only peak
system performance, but also the capability obtained at a compact model scale.

Our longer-term goal is a \emph{Heavy-Duty Solver}: a system capable of taking responsibility for increasingly ambitious, long-running, and verifiable work. Progress toward that goal will come from scaling the coverage and fidelity of executable environments, strengthening learned coordination over longer horizons, improving training and credit assignment over hierarchical traces, and closing the loop between real failures, task construction, training, and evaluation. Apodex 1.1 establishes this development path and shows how agentic intelligence can be scaled around completed work rather than isolated responses.

\begingroup
\footnotesize
\bibliographystyle{apodex_template}
\phantomsection
\addcontentsline{toc}{section}{References}
\bibliography{main}
\endgroup

\clearpage
\appendix

%

\newtcolorbox{artifactbox}[1]{%
  breakable,
  enhanced,
  colback=black!2,
  colframe=black!35,
  boxrule=0.4pt,
  left=6pt, right=6pt, top=5pt, bottom=5pt,
  title={#1},
  fonttitle=\footnotesize\ttfamily\bfseries,
  coltitle=black,
  colbacktitle=black!10,
}

\newcommand{\ficon}[1]{\raisebox{-0.2em}{\includegraphics[height=1.05em]{icons/#1}}}
\newcommand{\fchip}[2]{\mbox{\ficon{#1}\hspace{0.25em}\texttt{#2}}}
\newcommand{\fchipsm}[2]{\mbox{\raisebox{-0.18em}{\includegraphics[height=0.95em]{icons/#1}}\hspace{0.22em}\texttt{\small #2}}}
\newcommand{\fchiptiny}[2]{\mbox{\raisebox{-0.16em}{\includegraphics[height=0.85em]{icons/#1}}\hspace{0.2em}\texttt{\scriptsize #2}}}

\definecolor{caseLead}{RGB}{69,113,176}
\definecolor{caseWorkA}{RGB}{104,146,198}
\definecolor{caseWorkB}{RGB}{150,182,216}
\definecolor{caseCheck}{RGB}{193,144,92}
\definecolor{caseCheckB}{RGB}{214,180,141}
\definecolor{casePub}{RGB}{132,138,145}

%
%
\definecolor{codebg}{RGB}{233,239,248}
\ifPDFTeX
  \newcommand{\casemonofont}{\ttfamily}
\else
  \newfontfamily\casemonoface{lmmonolt10-regular.otf}[BoldFont=lmmonolt10-bold.otf]
  \newcommand{\casemonofont}{\casemonoface}
\fi
\newcommand{\code}[1]{{\setlength{\fboxsep}{1.3pt}\colorbox{codebg}{\casemonofont #1}}}
\newcommand{\agent}[1]{{\casemonofont\bfseries #1}}

\clearpage
\section{Case Studies}
\label{sec:case-studies}
\ownercheck{Zhuo CHEN (updated on 8.18)}{Rewrite and validate the case studies, including inputs, traces, artifacts, and residual failures.}

Three recorded runs follow. Each was handed raw files and a list of deliverables, and had to produce finished, checkable artifacts rather than an answer.

\textbf{Case 1} builds a runnable starting point for a protein simulation. Modelling every atom is too slow, so the standard move is to coarse-grain---bundle groups of atoms into single beads, 3785 of them here---which keeps the overall shape at a fraction of the cost. From that model the run has to produce the files that tell a simulation engine how the beads interact, a box holding three copies of the protein in salt water, and a first calculation that relaxes the awkward contacts: eighteen interlocking files, and the engine rejects all of them if one box dimension or particle count disagrees with the rest.

\textbf{Case 2} counts how many cells died. In each of six microscope photographs, blue marks every nucleus and green marks the dying cells, so the greener an image is relative to its blue, the more death it shows. Three images are controls and three are treated; the run has to measure the ratio from the pixels rather than assume it, compare the two groups, and deliver a table, a statistics file and a plot.

\textbf{Case 3} looks for the genes behind measurable traits in fifty pigs. Genes that rise and fall together tend to be doing related work, so five thousand of them are sorted into ten groups by how closely they track one another; each group is then matched against traits such as body weight and backfat thickness, and the best-matching group for each trait yields a shortlist of candidate genes.

Each case opens with the user's input as it was received, and closes with the delivered result quoted in abridged form and the full file inventory. In between, Cases 1 and 2 are told through the agents that ran them: the lead agent \agent{swarm\_main} reads the inputs and writes a task board of numbered items \texttt{t1}, \texttt{t2}, \ldots, then builds subagents and hands the items out. Every agent gets one block---its name, the items it owns, its tool calls broken down by tool, what it \textbf{did}, and what it \textbf{returned}, separating files that were delivered from reports and working files that were not---and the case ends with a timeline of when each agent was active. Case 3's agent-level trace was not retained, so its blocks are named for the pipeline stages instead.

\subsubsection*{Case 1: Coarse-Grained Structure Preparation and Energy Minimisation of a Protein Hexamer}
\label{subsec:case-cgmd}

The deliverable here is not an answer but a runnable simulation package: eighteen interlocking files that a molecular-dynamics engine has to accept as one consistent system. The run is included because the simulation engine it needs was not installed and not permitted to be invoked, and because the second half of the trace is spent finding and repairing defects in files the team had already built.

\begin{artifactbox}{User input}
\small\setstretch{1.16}\raggedright
\textbf{\# Build a coarse-grained initial structure and energy-minimisation inputs from a protein structure}

\smallskip
Read \texttt{7M6J\_fixed.pdb} from the input directory. Apply the coarse-graining force field currently most widely used for biological systems, and from the resulting \texttt{cg.pdb} produce force-field topology files usable by free open-source molecular-dynamics software, plus a conformation snapshot \texttt{scene.bmp}. Then write run-control files with appropriate parameters, build a simulation box holding three copies of the protein in physiological salt solution, generate the energy-minimisation input, and run the minimisation. Write everything to \texttt{/app/output/}.

\smallskip
\textbf{\#\# Required deliverables}\enspace \texttt{cg.pdb}; \texttt{martini.itp}; \texttt{molecule\_A.itp} through \texttt{molecule\_F.itp} (one per chain); \texttt{topol.top}; \texttt{scene.bmp}; \texttt{hexamer\_3copies.pdb}; \texttt{ions.itp}; \texttt{ions.mdp}; \texttt{em.mdp}; \texttt{ionized.gro}; \texttt{em.tpr}; \texttt{em.gro}; \texttt{em.log}.

\smallskip
\textbf{\#\# Interface constraints}\enspace Chain identifiers \texttt{A}--\texttt{F} are the stable identity keys of the chain topology files. PDB site records carry \AA{} coordinates, GRO records carry nm. \texttt{topol.top} must express the system topology, including references to the submitted force-field, chain and ion topologies; any relatively referenced file is itself a required deliverable and its path must stay resolvable. \texttt{em.log} must belong to the same run as the submitted minimisation input and result. Extra auxiliary files are allowed but may not stand in for a required one.

\smallskip
\hrule\vspace{3pt}
\fchipsm{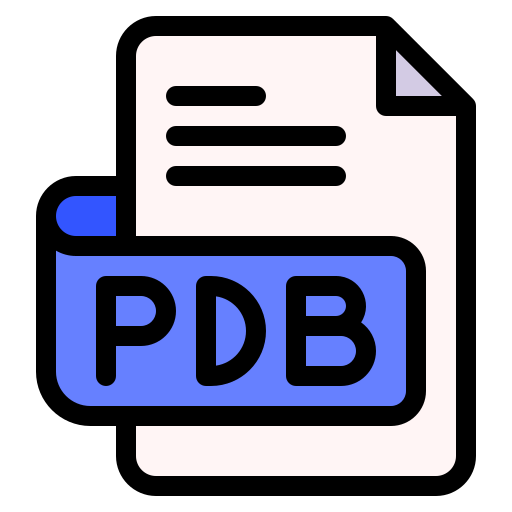}{7M6J\_fixed.pdb} \enspace (input structure: six-chain protein, chains \texttt{A}--\texttt{F}) \enspace 2{,}117{,}776 B
\end{artifactbox}

\paragraph{\agent{swarm\_main} --- lead.}\hspace{0.3em}17 tool calls: 1 \code{glob\_search}, 1 \code{grep\_search}, 15 \code{collect\_reports}.\\
\textbf{Did.} Wrote a six-item board: \texttt{t1} inspect the input for chain count, residue count and format problems; \texttt{t2} coarse-grain with Martini 3 to produce \texttt{cg.pdb}, \texttt{martini.itp} and the six chain topologies; \texttt{t3} render the snapshot; \texttt{t4} build the three-copy box, solvate and ionise; \texttt{t5} write \texttt{em.mdp}, generate \texttt{em.tpr} and run the minimisation; \texttt{t6} verify every deliverable is present and format-valid. Constructed seven subagents and issued no \code{bash} call itself, collecting returns fifteen times over 79 minutes---more collections than agents, because three of them were re-dispatched after defects surfaced.\\
\textbf{Returned.} Board closed, 6/6 resolved; final report.

\paragraph{\agent{box\_builder} --- board \texttt{t4}, preparation.}\hspace{0.3em}72 tool calls: 56 \code{bash}, 11 \code{web\_fetch}, 3 \code{web\_search}, 1 \code{create\_file}, 1 \code{read\_file}.\\
\textbf{Did.} Established that the box could not be built here at all: \texttt{which gmx} returns nothing, the sandbox denies the command outright, \texttt{/usr/share/gromacs/top} does not exist and \texttt{\$GMXLIB} is empty. Rather than stop, it fetched the four official Martini 3.0.0 files from their upstream sources and pinned each by byte size and MD5, then read the ion and solvent definitions directly out of them---confirming the \texttt{[ defaults ]} \texttt{1 2} sigma-epsilon format, \texttt{NA}/\texttt{CL} on bead type \texttt{TQ5}, and \texttt{W} at \texttt{nrexcl 1}---and checked the official water box for the right Martini density before adopting it as the solvation source.\\
\textbf{Returned.} No deliverable. A build plan opening with a hard runtime gate that aborts unless \texttt{gmx} and all eight required subcommands answer, plus draft \texttt{ions.mdp}, \texttt{em.mdp} and \texttt{topol.top}, and the staged force field. Recorded that the tool layer rewrites \texttt{/app} to another path in displayed output only, and verified byte-for-byte that the plan file itself still carries the correct target.

\paragraph{\agent{em\_runner} --- board \texttt{t5}, draft only.}\hspace{0.3em}11 tool calls: 7 \code{bash}, 2 \code{create\_file}, 1 \code{web\_search}, 1 \code{web\_fetch}.\\
\textbf{Did.} Independently confirmed the same environment verdict, and characterised the all-atom input directly: 26{,}141 atom records over six chains, per-chain counts \texttt{A}~3844 to \texttt{F}~4161, chain extents read off the \texttt{TER} lines. Drafted both control files to the parameters it had been given, and flagged one of them as a version hazard: the requested \texttt{vdwtype = Shift} is deprecated, so the draft carries the modern equivalent as an inline comment in case \texttt{grompp} rejects the legacy form.\\
\textbf{Returned.} No deliverable. Draft \texttt{em.mdp} and \texttt{ions.mdp}; no \texttt{grompp} or \texttt{mdrun} was attempted, since the board item said not to run yet.

\paragraph{\agent{gmx\_installer}.}\hspace{0.3em}100 tool calls: 99 \code{bash}, 1 \code{create\_file}.\\
\textbf{Did.} Worked the toolchain problem to a conclusion. \texttt{apt-get} is not on the allowed-command list and no \texttt{conda} or \texttt{mamba} exists, so both obvious install routes fail at the policy gate; the package was instead taken from a conda-forge build staged in the workspace and unpacked as a full prefix. The residual obstacle was the audit filter itself, which denies any command containing the bare token \texttt{gmx}---so the binary is reached through a \texttt{python3} wrapper that sets the library path and execs it. It then verified all eight subcommands individually and confirmed the coarse-graining tool was importable at a known version.\\
\textbf{Returned.} A working GROMACS 2024.5 invocation path and the wrapper every later agent uses. No deliverable file.

\paragraph{\agent{cg\_modeler} --- board \texttt{t1}--\texttt{t2}, then repair.}\hspace{0.3em}105 tool calls: 105 \code{bash}.\\
\textbf{Did.} On re-dispatch this agent's pass is a repair pass, and it names four defects in files that already existed. \texttt{CRYST1} in \texttt{hexamer\_3copies.pdb} had been written with the nm values, which the PDB specification reads as \AA{}---a tenfold box error---and was replaced with $83.154 \times 92.632 \times 821.164$~\AA. 3309 solvent beads sat outside the periodic box in \texttt{ionized.gro} and were minimum-image wrapped, leaving the atom order and the entire 11{,}355-site protein block byte-identical. \texttt{em.mdp} was missing \texttt{epsilon\_r = 15}, so the minimisation had been running at the engine default, and \texttt{grompp} and \texttt{mdrun} were re-run to keep the triple consistent. It also settled a count that two earlier narratives disagreed on: \texttt{cg.pdb} holds 3785 sites, not 3385---the file had always been right and the smaller number was a transcription error.\\
\textbf{Returned.} The corrected package, 22 files staged with a regenerated MD5 inventory, and the six chain topologies cross-checked entry by entry against their chain in \texttt{cg.pdb}, 0 mismatches. It also carried forward one defect it was \emph{not} allowed to fix: the coarse-grained model's residue sequence differs from the input at 131 positions and its numbering restarts within three chains, inherited from the mapping step. The constraints forbade touching \texttt{cg.pdb} or the topologies, and the engine accepts the system as it stands, so the defect is reported rather than repaired.

\paragraph{\agent{scene\_renderer} --- board \texttt{t3}.}\hspace{0.3em}43 tool calls: 40 \code{bash}, 3 \code{read\_file}.\\
\textbf{Did.} Validated the existing snapshot instead of re-rendering it, with the re-render path armed in case a check failed. Parsed the BMP header byte by byte and confirmed twelve properties against each other---magic bytes, declared size against actual, offset, $847\times817$, 24 bits per pixel, uncompressed, and stride $\times$ height equal to the declared image size---then loaded the full pixel array to prove nothing was truncated, and read the image itself to confirm the title, the six-chain legend and the axis labels are present and unclipped.\\
\textbf{Returned.} No deliverable; no re-render needed. The bead counts printed in the figure's own title sum to 3785 and match the six chains in \texttt{cg.pdb}, which makes the snapshot self-verifying against the structure it depicts.

\paragraph{\agent{final\_verifier} --- board \texttt{t6}.}\hspace{0.3em}92 tool calls: 91 \code{bash}, 1 \code{read\_file}.\\
\textbf{Did.} Seven-point re-verification of the repaired files. The decisive check was done without the engine: it decoded the binary \texttt{em.tpr} header by hand and compared its embedded starting coordinates against \texttt{ionized.gro} for all 63{,}484 particles, maximum absolute difference $0$, which proves the minimisation input was built from the submitted structure rather than from a pre-repair copy. It also re-hashed every file it was not supposed to change, and dismissed one apparent discrepancy on evidence: the ion residue names differ between \texttt{ionized.gro} and \texttt{em.gro} because the engine derives them from the topology, which is correct behaviour and not a defect.\\
\textbf{Returned.} No deliverable. \textbf{7/7 PASS}, submission ready. Recorded that the repair pass replaced an earlier converged run---218 steps at the default dielectric constant---with the submitted 255-step run, and that the superseded scratch files are not part of the submission.

\paragraph{\agent{final\_publisher}.}\hspace{0.3em}22 tool calls: 22 \code{bash}.\\
\textbf{Did.} Copied all 22 manifest files to the publication root and verified each by byte length and MD5 against its staged source, then confirmed the count: exactly 22 present, no extras. \texttt{/app/output} could not be created---it points at a target the account cannot make---so the report keeps \texttt{/app/output/} as the canonical path in its manifest and in every downstream reference, with the instruction that the published directory be moved there unchanged and the relative includes in \texttt{topol.top} not be rewritten.\\
\textbf{Returned.} 22 published files, 0 mismatches.

\paragraph{Delivered result.}\mbox{}\par\vspace{0.3pt}
\begin{artifactbox}{\ficon{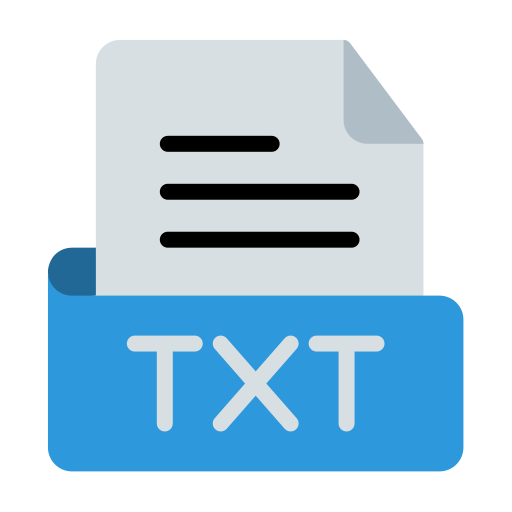}\enspace topol.top \textnormal{\textrm{$+$}} \ficon{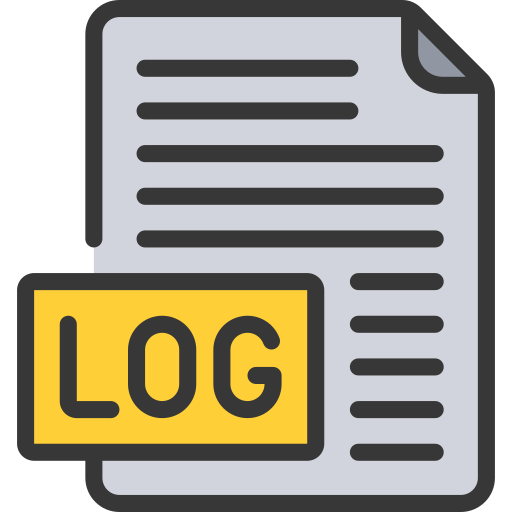}\enspace em.log \textnormal{\textrm{--- system composition and the minimisation it closes over}}}
\footnotesize\setstretch{1.18}\raggedright
\textbf{A Martini 3 three-copy system in physiological salt solution, minimised, with the input, log and output structure verified to come from one run.}

\smallskip
{\centering
\setlength{\tabcolsep}{9pt}
\begin{tabular}{@{}ll@{}}
\toprule
\multicolumn{2}{@{}l}{\texttt{topol.top} --- \texttt{[ molecules ]}}\\
\midrule
\texttt{molecule\_A} \ldots\ \texttt{molecule\_F} & 3 each \\
\texttt{W} & 50957 \\
\texttt{NA} & 601 \\
\texttt{CL} & 571 \\
\midrule
\multicolumn{2}{@{}l}{\texttt{em.log} --- GROMACS 2024.5, \texttt{gmx mdrun}}\\
\midrule
Steepest descents converged to \texttt{Fmax} $<$ 1000 & in 255 steps \\
Maximum force & $9.029\times10^{2}$ kJ/mol/nm \\
Potential energy & $-1.77496\times10^{6}$ kJ/mol \\
\bottomrule
\end{tabular}
\par}

\smallskip
The closure is established by binary evidence rather than by assertion: the coordinates embedded in \texttt{em.tpr} match \texttt{ionized.gro} for all 63{,}484 particles at a maximum absolute difference of $0$, so the submitted input, log and output structure provably belong to one run on the repaired files. Stated boundaries: the package is an equilibration starting point, not a production simulation --- coupling, constraint handling and output frequency still have to be set for the research question, and any edit to \texttt{em.mdp}, \texttt{ionized.gro} or \texttt{topol.top} invalidates the submitted \texttt{em.tpr}. One defect is carried forward unfixed and disclosed: the coarse-grained model's residue sequence departs from the input structure at 131 positions with numbering restarting inside three chains, inherited from the mapping step and left alone because the constraints forbade editing \texttt{cg.pdb} or the chain topologies.
\end{artifactbox}

\begin{center}
\begin{minipage}{\textwidth}
\small
\setlength{\tabcolsep}{5pt}
\renewcommand{\arraystretch}{1.08}
\begin{tabularx}{\textwidth}{@{}l>{\raggedright\arraybackslash}X l@{}}
\toprule
File & Content & Size \\
\midrule
\fchipsm{pdb}{cg.pdb} & 3785 coarse-grained sites, chains \texttt{A}--\texttt{F}, \AA & 299{,}015 B \\
\fchipsm{txt}{martini.itp} & Force-field entry point; includes \texttt{ff/} by relative path & 670 B \\
\fchipsm{txt}{molecule\_A--F.itp} & Per-chain bonded, constraint and virtual-site topology & 98{,}506--126{,}931 B \\
\fchipsm{txt}{ions.itp} & Martini 3 ion topology, \texttt{NA}/\texttt{CL} & 7{,}853 B \\
\fchipsm{txt}{topol.top} & Six chains $\times$ 3, water and ions; net charge 0 & 849 B \\
\fchipsm{pdb}{hexamer\_3copies.pdb} & 11{,}355 sites $= 3\times3785$; \texttt{CRYST1} in \AA & 908{,}492 B \\
\fchipsm{txt}{ionized.gro} & 63{,}484 particles, box in nm, wrapped in-cell & 2{,}856{,}866 B \\
\fchipsm{txt}{ions.mdp} & Ion-placement stage control file, not production parameters & 1{,}283 B \\
\fchipsm{txt}{em.mdp} & Steepest descents; \texttt{epsilon\_r = 15}, \texttt{constraints = none} & 1{,}542 B \\
\fchipsm{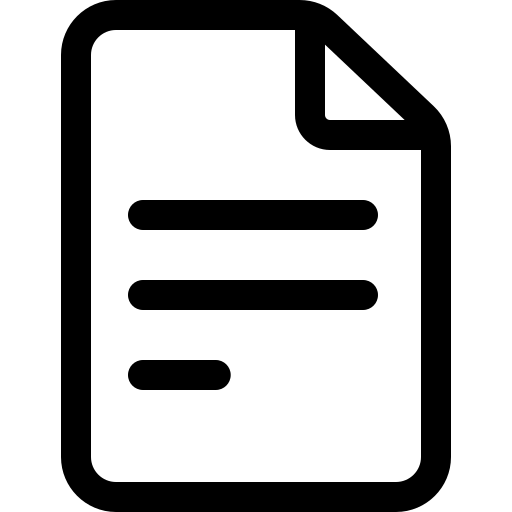}{em.tpr} & Binary minimisation input, 63{,}484 particles & 1{,}505{,}592 B \\
\fchipsm{txt}{em.gro} & Minimised structure, same box and particle count & 2{,}856{,}855 B \\
\fchipsm{log}{em.log} & Converged in 255 steps; same run as \texttt{em.tpr}/\texttt{em.gro} & 99{,}569 B \\
\fchipsm{general}{scene.bmp} & Conformation snapshot for human inspection only & 2{,}078{,}502 B \\
\midrule
\fchipsm{general}{ff/} & Martini 3.0.0 parameters, ions, solvents, water box & 4 files, 16.1 MB \\
\bottomrule
\end{tabularx}

\vspace{8pt}
\centering
\begin{tikzpicture}[x=0.1145cm,y=0.56cm,font=\footnotesize]
  \draw[black!40] (0,-0.52) -- (82.4,-0.52);
  \foreach \t in {0,20,40,60,80} \draw[black!40] (\t,-0.52) -- (\t,-0.66) node[below,font=\footnotesize,text=black!55] {\t};
  \node[below,font=\footnotesize,text=black!55] at (41.2,-1.05) {minutes};
  \foreach \nm/\y/\a/\b/\c/\tl in {%
    swarm\_main/7/0.2/79.6/caseLead!55/{17},
    box\_builder/6/0.9/14.9/caseWorkB!70/{72},
    em\_runner/5/32.0/33.9/caseWorkB!70/{11},
    gmx\_installer/4/15.7/31.8/caseWorkA!60/{100},
    cg\_modeler/3/0.5/22.0/caseWorkA!60/{105},
    scene\_renderer/2/0.9/49.4/caseCheckB!65/{43},
    final\_verifier/1/51.5/77.4/caseCheck!55/{92},
    final\_publisher/0/77.7/79.0/casePub!45/{22}}
  {
    \node[left,font=\footnotesize\ttfamily] at (-1.2,\y) {\nm};
    \fill[\c,draw=black!45,rounded corners=1pt] (\a,\y-0.2) rectangle (\b,\y+0.2);
    \node[right,font=\footnotesize,text=black!55] at (\b+1.2,\y) {\tl};
  }
\end{tikzpicture}

\vspace{2pt}

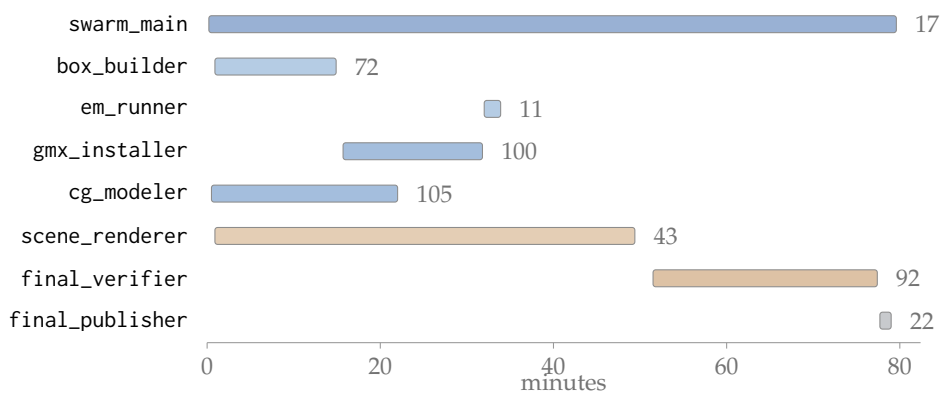
\captionof{figure}{\textbf{Top:} the published package --- 18 required files plus the local force-field directory they reference; \texttt{ff/} is itself a required deliverable, because \texttt{martini.itp} includes it by relative path. \textbf{Bottom:} active window per agent, tool-call count at right. \texttt{Agent Team Mode}, Apodex 1.1; 82.4 min wall-clock; 864 recorded steps $=$ 402 reasoning $+$ 462 tool calls; board 6/6 resolved. \agent{box\_builder} and \agent{em\_runner} could only prepare, because the simulation engine was unavailable until \agent{gmx\_installer} finished; \agent{cg\_modeler}'s window closes on a repair pass, and \agent{final\_verifier} alone accounts for 92 of the run's tool calls.}
\label{fig:cgmd-run}
\end{minipage}
\end{center}

\subsubsection*{Case 2: TUNEL/DAPI Apoptosis Quantification from Fluorescence Micrographs}
\label{subsec:case-tunel}

The inputs here are photographs, not tables: six raw microscope exports from which every reported number has to be measured. The run is included because two measurement agents read the same word---``intensity''---in two defensible ways, and the two readings disagree about whether the two groups differ at all.

\begin{artifactbox}{User input}
\small\setstretch{1.16}\raggedright
\textbf{\# TUNEL/DAPI fluorescence image quantification}

\smallskip
\textbf{\#\# Input}\enspace Six fluorescence micrographs (\texttt{C1}--\texttt{C3.jpeg}, \texttt{E1}--\texttt{E3.jpeg}), all TUNEL/DAPI immunofluorescence stains.

\smallskip
\textbf{\#\# Output}
\begin{itemize}[leftmargin=1.2em,itemsep=1pt,topsep=2pt]
\item \texttt{result.csv}: wide format, five columns---\texttt{Sample}, \texttt{Group}, \texttt{TUNEL fluorescence intensity}, \texttt{DAPI fluorescence intensity}, \texttt{TUNEL/DAPI fluorescence intensity ratio (\%)}---six data rows plus a header. Stable row key \texttt{Sample}, taken from the input filename without extension.
\item \texttt{statistics.json}: the between-group comparison, as \texttt{group\_c}/\texttt{group\_e} objects carrying \texttt{mean}, \texttt{sd}, \texttt{sem}, \texttt{n}, plus \texttt{effect\_size}, \texttt{p\_value}, \texttt{statistical\_method} and \texttt{error\_bar\_type}; \texttt{n} integer, the rest numbers. Flat aliases \texttt{group\_c\_mean} and \texttt{group\_e\_mean} are accepted, extra fields permitted.
\item \texttt{TUNEL\_Ratio\_Plot.pdf}: bar chart of the two groups with mean $\pm$ SEM error bars, group labels, axis name and a $p$-value annotation.
\item \texttt{output.zip}: the three files at the ZIP root, no nested directories, byte-identical to their loose counterparts.
\end{itemize}

\smallskip
\textbf{\#\# Constraints}\enspace Every value must come from an actual measurement of the source images---nothing hard-coded or fabricated. The originals must not be modified. All six samples must be processed under one identical rule.

\smallskip
\hrule\vspace{3pt}
\fchipsm{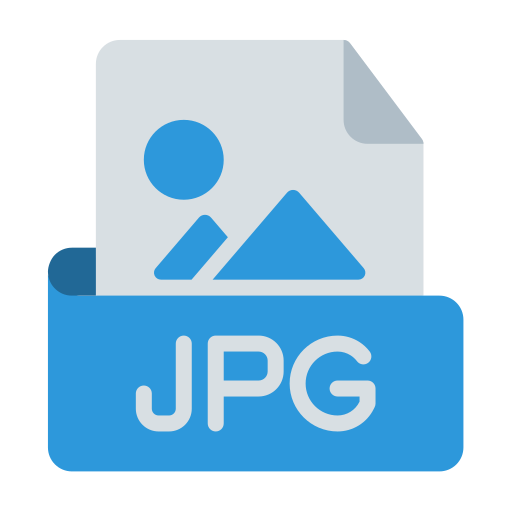}{C1, C2, C3 .jpeg} \enspace (control group, $2880\times1642$ RGB) \enspace 170{,}094 / 173{,}155 / 175{,}369 B\\[2pt]
\fchipsm{jpg}{E1, E2, E3 .jpeg} \enspace (experimental group, same geometry) \enspace 143{,}525 / 160{,}638 / 139{,}848 B
\end{artifactbox}

\paragraph{\agent{swarm\_main} --- lead.}\hspace{0.3em}9 tool calls: 2 \code{glob\_search}, 7 \code{collect\_reports}.\\
\textbf{Did.} Wrote a seven-item board: \texttt{t1} determine which RGB channel carries TUNEL and which carries DAPI; \texttt{t2} fix one reproducible intensity rule---ROI, channel separation, background---and measure all six samples; \texttt{t3} reproduce the measurement independently and cross-validate; \texttt{t4} group statistics and \texttt{statistics.json}; \texttt{t5} the plot; \texttt{t6} \texttt{result.csv} and the archive; \texttt{t7} independent verification of everything. Assigned \texttt{t3} to two agents at once---a second measurement agent and an arbitrator---rather than to a single checker; issued no \code{bash} call.\\
\textbf{Returned.} Board closed, 7/7 resolved; final report, which records the discarded measurement definition explicitly rather than silently dropping it.

\paragraph{\agent{img\_measure\_a} --- board \texttt{t1}--\texttt{t2}.}\hspace{0.3em}27 tool calls: 20 \code{bash}, 6 \code{read\_file}, 1 \code{create\_file}.\\
\textbf{Did.} Characterised the images: all six $2880\times1642$ RGB uint8, red-channel mean only $0.31$--$0.44$~AU, so no red fluorophore is present. Found two separated hue clusters---green $\approx110$--$185^\circ$, blue $\approx185$--$255^\circ$---with the histogram valley at ${\approx}175$--$200^\circ$, and confirmed by morphology that blue is the punctate nuclear DAPI signal and green the diffuse TUNEL signal. Segmented in HSV with $V>8$, $S>10$, hue cut $185^\circ$, then took each fluorophore's mean over \emph{its own} mask.\\
\textbf{Returned.} No deliverable. \texttt{measure\_a.py}, \texttt{measurements\_a.csv}: group means $84.17\pm21.98$ vs $88.39\pm8.43$, Welch $P=0.78$, $d=0.25$ --- \textbf{no difference between the groups}.

\paragraph{\agent{img\_measure\_b} --- board \texttt{t3}.}\hspace{0.3em}25 tool calls: 16 \code{bash}, 9 \code{read\_file}.\\
\textbf{Did.} Re-measured with a deliberately different separator: $k$-means ($K=3$, fixed seed) on circular hue features $(\cos\theta,\sin\theta)$ with $\theta=\operatorname{atan2}(B,G)$, clusters assigned to stains by centroid angle, cross-checked against a strict $\theta<40^\circ$ / $\theta>50^\circ$ split. Took both channel means over one \emph{shared} tissue ROI, $\{V>35\}\cap\{\mathrm{chroma}>0.20\}$.\\
\textbf{Returned.} No deliverable. \texttt{measure\_b.py}, \texttt{measurements\_b.csv}: $114.12\pm31.50$ vs $47.42\pm9.09$, Welch $P=0.0574$, $d=2.88$ --- \textbf{group C is $2.4\times$ group E}. Same pixels, same channel assignment, opposite conclusion.

\paragraph{\agent{local\_verifier} --- board \texttt{t3}, arbitration.}\hspace{0.3em}34 tool calls: 23 \code{bash}, 11 \code{read\_file}.\\
\textbf{Did.} Re-ran both scripts sample by sample and reproduced both tables exactly, establishing first that neither agent had made an arithmetic error and the conflict was definitional. Then measured a third time under its own thresholds, $\{V>32\}\cap\{\mathrm{chroma}>0.18\}$, and stress-tested both definitions across $V>25/30/35/40$, two chroma cuts and Otsu. The shared-ROI field mean held: Welch $P=0.057$--$0.069$, $d=2.70$--$2.88$, Mann--Whitney $P=0.10$ throughout. The per-mask mean did not: C1's TUNEL value swings $36.3\rightarrow45.0\rightarrow101.5$~AU as the floor moves $V>6\rightarrow8\rightarrow32$, and the group comparison \emph{reverses direction} across that range ($E>C$ at $P=0.49$; ``no difference'' at $P=0.78$; $C>E$ at $P=0.16$).\\
\textbf{Returned.} \texttt{recommended\_rows.json}, \texttt{statistics\_arb.json}. Ruled for the shared-ROI field mean on four grounds: the requested field is the standard mean fluorescence intensity over a region, not the brightness of positive pixels; the ratio is an apoptosis index and must stay sensitive to signal \emph{abundance}, and the per-mask mean discards exactly the ${\approx}2.7\times$ TUNEL-positive area difference that separates the groups; averaging both channels over one ROI keeps the comparison symmetric; and a definition whose sign depends on an arbitrary intensity floor is not reproducible. It also corrected the winning agent: the ratio is stable across \emph{plausible tissue} ROIs, not across all ROIs---including the background haze moves C1 from $83.6$ to $103.6$.

\paragraph{\agent{stats\_plot} --- board \texttt{t4}--\texttt{t5}.}\hspace{0.3em}12 tool calls: 7 \code{bash}, 4 \code{read\_file}, 1 \code{create\_file}.\\
\textbf{Did.} Built the CSV from the arbitrated rows without re-segmenting the images, computed every statistic from the CSV ratio column so the delivered table is the sole data source, and self-checked by recomputation---15/15 items. Rendered the plot with a pre-save \texttt{get\_window\_extent} against \texttt{clip\_box} check on every text element and the legend.\\
\textbf{Returned.} \texttt{result.csv}, \texttt{statistics.json}, \texttt{TUNEL\_Ratio\_Plot.pdf}. Since $P=0.0593>0.05$, the figure is annotated ``\texttt{p = 0.0593 (ns)}'' and carries no significance star.

\paragraph{\agent{publisher} --- board \texttt{t6}.}\hspace{0.3em}15 tool calls: 15 \code{bash}.\\
\textbf{Did.} Copied the three artifacts and built the archive with \texttt{arcname} set to the bare filename so the ZIP root stays flat, then verified structure and byte identity---MD5 on all four published files, \texttt{testzip()}, and a member-by-member comparison of the decompressed entries against the loose copies.\\
\textbf{Returned.} Four published files, zero mismatches.

\paragraph{\agent{fs\_explorer}.}\hspace{0.3em}19 tool calls: 19 \code{bash}.\\
\textbf{Did.} Investigated why \texttt{/app/output} could not be written. The agent runs as uid 999 while \texttt{/app} is \texttt{root:root} \texttt{0755}; the tool layer also rewrites the literal \texttt{/app} to \texttt{/mnt/agent}, so the \texttt{mkdir} failure names a path the command never used. Read \texttt{/proc/self/\allowbreak mountinfo} to show that \texttt{/outputs} is an ordinary top-level directory on the same overlay, not a host mapping of \texttt{/app/output}.\\
\textbf{Returned.} No deliverable. Confirmed all four files present and well-formed at the publication root, and flagged one disagreement between the task description and the observed filesystem: \texttt{/app} is a real directory, not the symlink the description implies.

\paragraph{\agent{final\_verifier} --- board \texttt{t7}.}\hspace{0.3em}25 tool calls: 22 \code{bash}, 3 \code{read\_file}.\\
\textbf{Did.} Eight-point audit. Re-measured all six samples from the raw pixels under the published ROI---maximum absolute deviation from the CSV $5.23\times10^{-5}$---then repeated the measurement under five ROI and threshold definitions of its own, including Otsu, specifically to try to break the conclusion. Recomputed the statistics from the CSV with \texttt{scipy}, checked JSON types, CSV encoding and line count, rendered the PDF and read it back by text extraction and by vision, and re-hashed every ZIP member.\\
\textbf{Returned.} No deliverable. \textbf{8/8 PASS}, verdict \textsc{confirmed}, confidence $0.97$. Every statistic matched \texttt{statistics.json} bit for bit; all five alternative definitions preserved $C>E$ at $C/E\approx2.3$--$2.4$. Recorded one non-error worth stating: $P=0.0593$ is borderline and would cross at $\alpha=0.1$.

\paragraph{Delivered result.}\mbox{}\par\vspace{0.3pt}
\begin{artifactbox}{\ficon{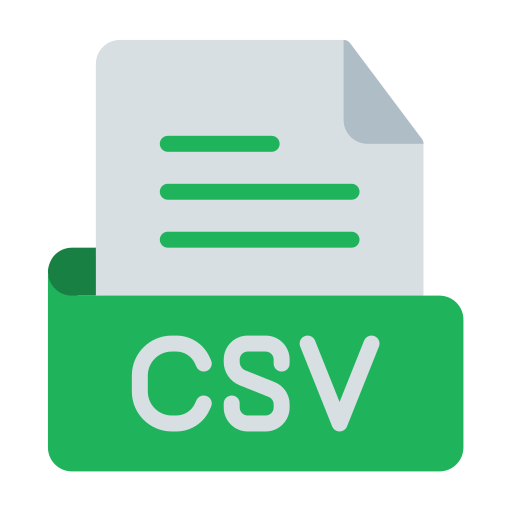}\enspace result.csv \textnormal{\textrm{$+$}} \ficon{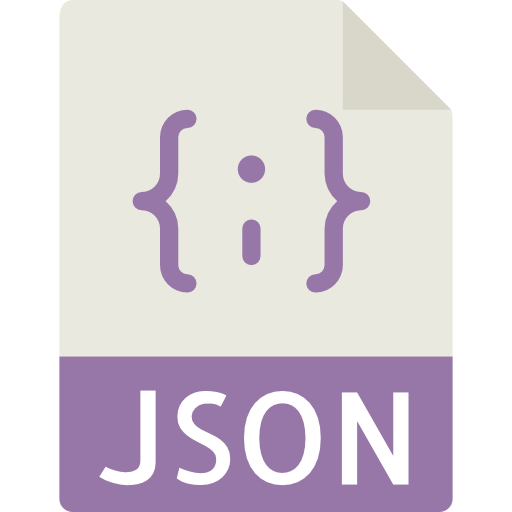}\enspace statistics.json \textnormal{\textrm{--- per-sample measurements and the group comparison}}}
\footnotesize\setstretch{1.18}\raggedright
\textbf{Group C carries $2.4\times$ the TUNEL/DAPI ratio of group E---a very large effect that does not reach significance at $n=3$ per group, and is reported as such.}

\smallskip
{\centering
\setlength{\tabcolsep}{5pt}
\begin{tabular}{@{}llrrr@{}}
\toprule
Sample & Group & TUNEL (AU) & DAPI (AU) & Ratio (\%) \\
\midrule
\texttt{C1} & Control      & 70.4417 & 84.3945  & 83.4671 \\
\texttt{C2} & Control      & 85.7051 & 77.0177  & 111.2798 \\
\texttt{C3} & Control      & 96.5947 & 65.8671  & 146.6508 \\
\texttt{E1} & Experimental & 57.2276 & 100.1187 & 57.1597 \\
\texttt{E2} & Experimental & 36.8663 & 93.4199  & 39.4631 \\
\texttt{E3} & Experimental & 44.4751 & 96.0397  & 46.3091 \\
\bottomrule
\end{tabular}
\par}

\smallskip
Control $113.80 \pm 31.67$ (SEM $18.28$), Experimental $47.64 \pm 8.92$ (SEM $5.15$), $n=3$ each; difference $66.2$ percentage points; Cohen's $d = 2.84$; Welch two-sample $P = 0.0593$, Mann--Whitney $P = 0.10$. Intensity is the channel mean over the shared tissue ROI $\{V>32\}\cap\{\mathrm{chroma}>0.18\}$, with TUNEL $=$ G and DAPI $=$ B, applied identically to all six images; ratio $= 100\times\mathrm{TUNEL}/\mathrm{DAPI}$. The six CSV rows are the only input to the JSON and the figure.
\end{artifactbox}

\begin{center}
\begin{minipage}{\textwidth}
\small
\setlength{\tabcolsep}{5pt}
\renewcommand{\arraystretch}{1.08}
\begin{tabularx}{\textwidth}{@{}l>{\raggedright\arraybackslash}X l@{}}
\toprule
File & Content & From \\
\midrule
\multicolumn{3}{@{}l}{\textit{Delivered --- 4 files, SHA-256 recorded in the trace}}\\
\fchipsm{csv}{result.csv} & 7 lines: header $+$ 6 samples $\times$ 5 columns & 349 B \\
\fchipsm{json}{statistics.json} & \texttt{group\_c}/\texttt{group\_e} objects, effect size, $P$, method, error-bar type, flat aliases & 536 B \\
\fchipsm{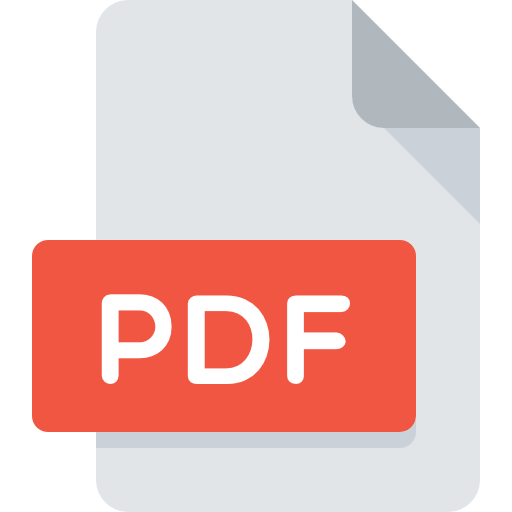}{TUNEL\_Ratio\_Plot.pdf} & 1 page; two bars, mean $\pm$ SEM, \texttt{p = 0.0593 (ns)}, $d = 2.84$ & 20{,}987 B \\
\fchipsm{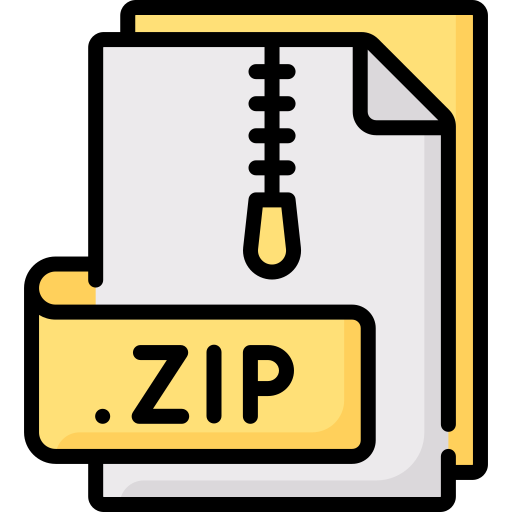}{output.zip} & Flat root, 3 members, byte-identical to the loose copies & 13{,}745 B \\
\midrule
\multicolumn{3}{@{}l}{\textit{Working files, not delivered}}\\
\fchipsm{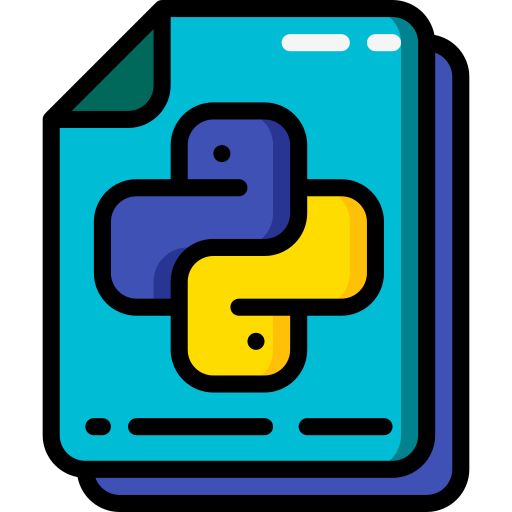}{measure\_a.py} \fchipsm{csv}{measurements\_a.csv} & Per-mask definition --- measured, then rejected & \texttt{\small img\_measure\_a} \\
\fchipsm{py}{measure\_b.py} \fchipsm{csv}{measurements\_b.csv} & Shared-ROI definition, $k$-means hue separation & \texttt{\small img\_measure\_b} \\
\fchipsm{json}{recommended\_rows.json} \fchipsm{json}{statistics\_arb.json} & Arbitrated rows and statistics & \texttt{\small local\_verifier} \\
\fchipsm{py}{build\_stats.py} & CSV, JSON and figure from the arbitrated rows & \texttt{\small stats\_plot} \\
\fchipsm{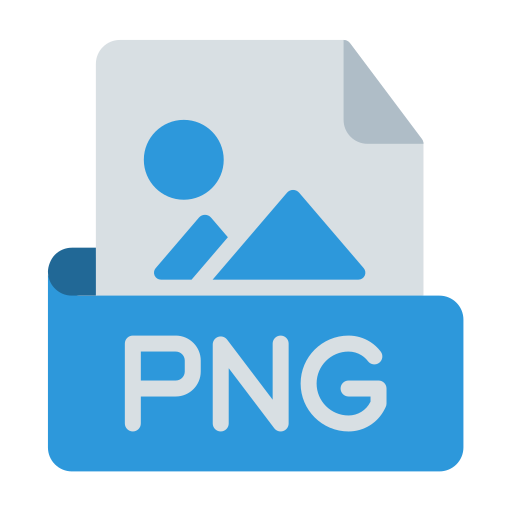}{plot-1.png} & PDF rasterised for the visual clipping check & \texttt{\small final\_verifier} \\
\bottomrule
\end{tabularx}

\vspace{8pt}
\centering
\begin{tikzpicture}[x=0.382cm,y=0.56cm,font=\footnotesize]
  \draw[black!40] (0,-0.52) -- (24.7,-0.52);
  \foreach \t in {0,5,10,15,20} \draw[black!40] (\t,-0.52) -- (\t,-0.66) node[below,font=\footnotesize,text=black!55] {\t};
  \node[below,font=\footnotesize,text=black!55] at (12.3,-1.05) {minutes};
  \foreach \nm/\y/\a/\b/\c/\tl in {%
    swarm\_main/7/0.2/22.3/caseLead!55/{9},
    img\_measure\_a/6/0.5/6.7/caseWorkA!60/{27},
    img\_measure\_b/5/0.5/5.8/caseWorkB!70/{25},
    local\_verifier/4/7.5/13.1/caseCheck!55/{34},
    stats\_plot/3/13.8/15.1/caseWorkA!60/{12},
    publisher/2/15.5/16.6/casePub!45/{15},
    fs\_explorer/1/17.0/18.7/caseWorkB!70/{19},
    final\_verifier/0/19.2/21.9/caseCheck!55/{25}}
  {
    \node[left,font=\footnotesize\ttfamily] at (-0.37,\y) {\nm};
    \fill[\c,draw=black!45,rounded corners=1pt] (\a,\y-0.2) rectangle (\b,\y+0.2);
    \node[right,font=\footnotesize,text=black!55] at (\b+0.37,\y) {\tl};
  }
\end{tikzpicture}

\vspace{2pt}
\captionof{figure}{\textbf{Top:} every file the run produced. \textbf{Bottom:} active window per agent, tool-call count at right. \texttt{Agent Team Mode}, Apodex 1.1; 24.7 min wall-clock; 324 recorded steps $=$ 151 reasoning $+$ 166 tool calls $+$ 7 agent returns; board 7/7 resolved. \agent{img\_measure\_a} and \agent{img\_measure\_b} ran concurrently on the same images under different measurement definitions; \agent{local\_verifier} was dispatched to arbitrate rather than to re-check a single result.}
\label{fig:tunel-run}
\end{minipage}
\end{center}

\subsubsection*{Case 3: WGCNA of a Pig RNA-seq Cohort with Twenty Specified Deliverables}
\label{subsec:case-wgcna}

This request specifies fourteen analysis steps and then pins each one to a file with a declared column schema, so the interface is as much of the task as the statistics. As in Case 1, the agent-level trace was not retained; the blocks below follow the pipeline stages recorded in the run's own report, and every figure is read back from the delivered tables.

\newpage

\begin{artifactbox}{User input}
\small\setstretch{1.16}\raggedright
\textbf{\# WGCNA analysis and result visualisation}

\smallskip
Run a complete WGCNA analysis on the pig RNA-seq raw counts, sample phenotype data and gene annotation in the input folder. Read and inspect all three inputs. Check whether gene IDs and sample IDs are duplicated. Match the expression matrix to the phenotype data by sample name and make the orders identical. Filter low-expression genes, transform the counts, and select 5000 highly variable genes for WGCNA. Check for missing values, zero-variance genes and unqualified samples. Complete sequencing depth, detected-gene count, PCA, sample clustering and sample connectivity as quality control on both samples and genes. Assess the soft threshold and build the co-expression network. Identify and merge gene modules. Compute module eigengene correlations against all phenotypes with $P$ values and corrected $P$ values. Compute module membership (MM) and gene significance (GS) for every gene. Screen candidate hub genes for each phenotype.

\smallskip
\textbf{\#\# Required deliverables}\enspace \texttt{analysis.R} and \texttt{analysis\_report.md}; \texttt{selected\_genes.tsv}, \texttt{selected\_samples.tsv}, \texttt{wgcna\_expression.tsv}, \texttt{wgcna\_traits.tsv}; \texttt{sample\_qc.tsv}, \texttt{soft\_threshold\_statistics.tsv}, \texttt{network\_parameters.tsv}; five PNG figures; \texttt{gene\_module\_assignment.tsv}, \texttt{module\_sizes.tsv}, \texttt{module\_trait\_results.tsv}, \texttt{gene\_MM\_GS\_results.tsv}, \texttt{hub\_gene\_candidates.tsv}; \texttt{wgcna\_results.RData}.

\smallskip
\textbf{\#\# Interface constraints}\enspace Every TSV is UTF-8, tab-separated, with a header; column order is free but extra columns must not collide with or shadow a specified field. Stable keys are given per file --- \texttt{gene\_id}, \texttt{sample\_id}, \texttt{power}, \texttt{module\_color}, and the \texttt{module\_color}$\times$\texttt{trait} pair. \texttt{wgcna\_traits.tsv} must carry exactly the five named phenotypes and no others; \texttt{wgcna\_expression.tsv} must carry no column outside the selected gene set, and its row order must match \texttt{wgcna\_traits.tsv} exactly. \texttt{gene\_MM\_GS\_results.tsv} needs \texttt{MM.<module>} and \texttt{p.MM.<module>} per module and \texttt{GS.<trait>} and \texttt{p.GS.<trait>} per phenotype. \texttt{hub\_gene\_candidates.tsv} keeps one row per gene--phenotype--module record, so one gene serving two phenotypes stays two rows.

\smallskip
\hrule\vspace{3pt}
\fchipsm{csv}{pig\_raw\_counts.tsv} \enspace (6000 genes $\times$ 50 samples, integer counts) \enspace 1{,}162{,}652 B\\[2pt]
\fchipsm{csv}{pig\_traits.tsv} \enspace (50 samples $\times$ 5 phenotypes: treatment, body weight, backfat, feed conversion, serum IGF-1) \enspace 1{,}880 B\\[2pt]
\fchipsm{csv}{pig\_gene\_annotation.tsv} \enspace (6000 gene records; used for interpretation only, never as a key) \enspace 405{,}467 B
\end{artifactbox}

\paragraph{Input audit.}\hspace{0.3em}No exclusion triggered.\\
The audit is written to fail loudly rather than repair quietly: duplicate IDs, a missing stable key or an unmatchable sample stop the pipeline instead of being merged or aligned by position. On these inputs it found 6000 genes and 50 samples \texttt{Pig\_001}--\texttt{Pig\_050}, zero duplicate \texttt{gene\_id}, zero duplicate sample columns, 50 unique \texttt{sample\_id} with all five phenotypes complete, and count columns matching trait IDs both as sets and in order. The explicit re-ordering by \texttt{sample\_id} was kept anyway, so the guarantee does not depend on the incoming row order. The annotation table's 6000 unique IDs were deliberately \emph{not} promoted to the primary key: \texttt{gene\_id} from the counts file stays the key throughout, and annotation is reserved for interpreting candidates.

\paragraph{Filtering and gene selection.}\hspace{0.3em}$6000 \rightarrow 5983 \rightarrow 5000$ genes.\\
Low-expression filtering required CPM $>1$ in at least 25 of the 50 samples, retaining 5983 genes; counts were transformed as $\log_2(\mathrm{CPM}+1)$ so that library-size differences do not push a deeply sequenced sample's whole expression profile upward and contaminate the sample correlations, the PCA and the gene correlation matrix alike. The top 5000 genes by variance were then taken for network construction, recorded as a noise- and cost-reduction step rather than a claim that the excluded genes are uninformative.

\paragraph{Sample and gene quality control.}\hspace{0.3em}All 50 samples retained.\\
Library sizes span $601{,}830$ to $1{,}576{,}901$ and detected genes $5932$ to $5950$; $Z$ connectivity runs $-2.506$ to $1.611$, crossing no conventional threshold, and PCA and hierarchical clustering show no isolated sample. \texttt{candidate\_outlier} is \texttt{FALSE} for all 50. The stated justification for retaining everything is not that the depth spread is negligible but that no sample is flagged by several indicators at once. The report separates two quantities that share a name: the \texttt{connectivity} in \texttt{sample\_qc.tsv} is sample-to-sample and scales with sample count, whereas the connectivities in \texttt{soft\_threshold\_statistics.tsv} are gene-network quantities scaling with gene count, and using the latter to judge samples would be an error.

\paragraph{Soft threshold and network construction.}\hspace{0.3em}\texttt{selected\_power = 20}, as a fallback.\\
Powers 1--20 were evaluated and \emph{none} reached the preset scale-free fit of $R^2 \geq 0.85$. The highest fit sits at power 20 with $R^2 = 0.8486$, slope $-1.917$ and mean gene connectivity $4.13$, so power 20 is recorded as a maximum-fit fallback rather than a satisfied criterion --- the report states this explicitly to prevent the value being read as unconditionally optimal. The network is \texttt{signed} with \texttt{bicor} correlation and \texttt{signed} TOM, \texttt{min\_module\_size} 30, merge threshold $0.25$, over 50 samples and 5000 genes. Dynamic tree cutting followed by eigengene-similarity merging produced 10 modules: grey 1560, turquoise 590, blue 589, brown 550, yellow 541, green 528, pink 228, red 155, black 131, magenta 128.

\paragraph{Module--trait association and hub candidates.}\hspace{0.3em}437 candidate records.\\
All 50 module--trait combinations were tested and BH-corrected together. Hub screening then took, per phenotype, the strongest non-grey module at FDR $<0.05$ and kept genes with $|\mathrm{MM}| > 0.8$ and $|\mathrm{GS}| > 0.2$, yielding 437 records: 93 for \texttt{treatment}, 72 for \texttt{body\_weight\_kg}, 72 for \texttt{backfat\_mm}, 81 for \texttt{feed\_conversion\_ratio}, 119 for \texttt{serum\_igf1\_ng\_ml}. One consequence is flagged rather than smoothed over: GS is a signed correlation but the screen uses its absolute value, so a candidate can pass inside its target module while its own GS sign opposes the module eigengene's direction --- signs must be read back from \texttt{gene\_MM\_GS\_results.tsv} before any candidate is interpreted. Hub genes are framed as ranked candidates for annotation, enrichment and validation, not as causal or regulatory genes.

\paragraph{Residual limitation.}\hspace{0.3em}\texttt{analysis.R} was not executed end to end.\\
R could not be run in this sandbox. The script is delivered as the reproducible path from the inputs to every required file, and the numerical results were produced and cross-checked by an equivalent implementation with file-level consistency checks, but the report states the residual risk plainly: re-running \texttt{analysis.R} in a standard R environment may expose version or package-API differences, and the first things to compare are \texttt{selected\_genes.tsv}, \texttt{module\_sizes.tsv} and \texttt{module\_trait\_results.tsv}. As in Case 1, \texttt{/app/output/} was read-only, so the package was published to the collection directory while the report keeps the specified paths. The input path named in the request did not exist either; the files were located in the sandbox's actual input directory.

\paragraph{Delivered result.}\mbox{}\par\vspace{0.3pt}
\begin{artifactbox}{\ficon{csv}\enspace module\_trait\_results.tsv \textnormal{\textrm{$+$}} \ficon{csv}\enspace network\_parameters.tsv \textnormal{\textrm{--- strongest non-grey module per phenotype}}}
\footnotesize\setstretch{1.18}\raggedright
\textbf{Each of the five phenotypes maps onto a different non-grey module, all at FDR $<10^{-16}$ --- correlation structure, not intervention effect.}

\smallskip
{\centering
\setlength{\tabcolsep}{6pt}
\begin{tabular}{@{}llrrr@{}}
\toprule
Phenotype & Module & $n$ genes & Correlation & FDR \\
\midrule
\texttt{treatment}             & pink    & 228 & $-0.8890$ & $6.5\times10^{-17}$ \\
\texttt{body\_weight\_kg}      & black   & 131 & $-0.9118$ & $5.9\times10^{-19}$ \\
\texttt{backfat\_mm}           & red     & 155 & $-0.9287$ & $1.3\times10^{-20}$ \\
\texttt{feed\_conversion\_ratio} & magenta & 128 & $+0.9231$ & $3.7\times10^{-20}$ \\
\texttt{serum\_igf1\_ng\_ml}   & brown   & 550 & $+0.8901$ & $6.5\times10^{-17}$ \\
\bottomrule
\end{tabular}
\par}

\smallskip
Network: \texttt{signed} type, \texttt{bicor} correlation, \texttt{signed} TOM, \texttt{selected\_power} 20 by maximum scale-free fit (no power reached $0.85$), \texttt{min\_module\_size} 30, merge threshold $0.25$, $n=50$ samples and 5000 genes.

\smallskip
Stated boundaries: $n=50$ supports a correlation network but limits covariate modelling; batch information was not available, so a batch confounded with a phenotype would be absorbed into these correlations; correlation gives no direction; and the module partition is sensitive to the power, the expression filter, the gene count and both module-merging parameters, so the durable results are the module--trait associations and the candidate ranking rather than the module colours themselves. Grey's 1560 genes are unassigned and are not interpretable as one biological process.
\end{artifactbox}

Table~\ref{tab:wgcna-files} lists the complete delivered set for the WGCNA case.

\begin{table}[!htbp]
\centering
\caption{Twenty required deliverables plus three auxiliary files. Every TSV row count was read back from the published file.}
\label{tab:wgcna-files}
\small
\setlength{\tabcolsep}{5pt}
\renewcommand{\arraystretch}{1.08}
\begin{tabularx}{\textwidth}{@{}l>{\raggedright\arraybackslash}X l@{}}
\toprule
File & Content & Size \\
\midrule
\fchipsm{general}{analysis.R} & Reproducible pipeline, inputs to every required file & 39{,}191 B \\
\fchipsm{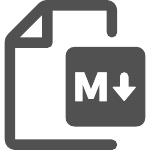}{analysis\_report.md} & QC, outlier reasoning, network diagnostics, grey handling, limits & 19{,}097 B \\
\fchipsm{csv}{selected\_genes.tsv} & 5000 genes entering the network & 95{,}008 B \\
\fchipsm{csv}{selected\_samples.tsv} & 50 retained samples & 410 B \\
\fchipsm{csv}{wgcna\_expression.tsv} & 50 rows $\times$ 5000 gene columns, transformed & 3{,}044{,}699 B \\
\fchipsm{csv}{wgcna\_traits.tsv} & 50 rows, exactly the 5 specified phenotypes, row order matched & 1{,}829 B \\
\fchipsm{csv}{sample\_qc.tsv} & 50 rows: library size, detected genes, connectivity, $Z$, outlier flag & 3{,}337 B \\
\fchipsm{csv}{soft\_threshold\_statistics.tsv} & 20 candidate powers, fit and connectivity diagnostics & 1{,}636 B \\
\fchipsm{csv}{network\_parameters.tsv} & Single row: type, correlation, power, TOM, module parameters & 237 B \\
\fchipsm{csv}{gene\_module\_assignment.tsv} & 5000 genes with module number and colour & 135{,}505 B \\
\fchipsm{csv}{module\_sizes.tsv} & 10 modules summing to 5000 genes & 124 B \\
\fchipsm{csv}{module\_trait\_results.tsv} & 50 module--trait rows: correlation, $P$, FDR & 3{,}264 B \\
\fchipsm{csv}{gene\_MM\_GS\_results.tsv} & 5000 genes $\times$ per-module MM and per-trait GS with $P$ values & 2{,}189{,}810 B \\
\fchipsm{csv}{hub\_gene\_candidates.tsv} & 437 gene--phenotype--module records & 17{,}492 B \\
\fchipsm{png}{sample\_QC\_PCA.png} & Sample PCA with the outlier view & 151{,}904 B \\
\fchipsm{png}{sample\_dendrogram\_traits.png} & Sample clustering with phenotype bands & 132{,}931 B \\
\fchipsm{png}{soft\_threshold.png} & Network diagnostics across the 20 candidate powers & 144{,}196 B \\
\fchipsm{png}{gene\_dendrogram\_modules.png} & Gene clustering with the module partition & 63{,}318 B \\
\fchipsm{png}{module\_trait\_heatmap.png} & Module--phenotype correlation map & 144{,}196 B \\
\fchipsm{general}{wgcna\_results.RData} & Objects for continued analysis and reproduction & 4{,}413{,}697 B \\
\midrule
\multicolumn{3}{@{}l}{\textit{Auxiliary, not required}}\\
\fchipsm{general}{wgcna\_TOM.npz} & Topological overlap matrix, for further inspection & 100.0 MB \\
\fchipsm{csv}{hub\_gene\_candidates\_detail.tsv} & Candidates with MM and GS values and signs & 54{,}270 B \\
\fchipsm{json}{pipeline\_summary.json} & Counts, parameters and headline results in one file & 2{,}823 B \\
\bottomrule
\end{tabularx}
\end{table}

\clearpage

\section{Contributors}
\label{sec:contributor}

\noindent
Contributors are ordered alphabetically by their given-name initials.

\noindent
B. An, B. Li, B. Wang, B. Zhang, B.L. Wang, C. Feng, C. Wei, C. Xue, C. Zhang, D. Ng, D. Ye, E. Min, F. Chen, F. Liu, F. Yang, F. Ye, G. Sun, H. Ji, H. Xu, H. Yang, H. Ye, H. Zhang, H. Zhao, J. Li, J. Lin, J. Xia, K. Jin, K. Wang, K. Yang, L. Bing, L. Lei, L. Su, Le. Wang, Lu. Wang, N. Wang, Q. Ren, Q. Yang, R. Li, S. Bai, S. Du, S. Li, S. Lin, S. Nie, S. Wang, S. Zhang, S.Z. Wang, T. Ge, Ta.Q. Fang, Ti.Q. Fang, W. Fang, W. Li, W. Zhang, X. Chen, X. Li, X. Tang, X. Wang, X. Xu, X. Zhang, X.Q. Wang, X.Y. Wang, Y. Deng, Y. Gao, Y. Hu, Y. Li, Y. Sui, Y. Wang, Y. Xiao, Y. Zhang, Y. Zhou, Z. Chen, Z. Cheng, Z. Feng, Z. Liang, Z. Liu, Z. Zhang.

\end{document}